\documentclass[11pt,a4paper,onecolumn]{article}
\usepackage[left=2.5cm,right=2.5cm,top=2cm,bottom=2cm]{geometry}

\usepackage{graphicx}
\usepackage{amsmath}
\usepackage{amsfonts}
\usepackage{amssymb}
\usepackage{subfigure}
\usepackage{algorithm}
\usepackage{algorithmic}
\usepackage{dsfont}
\usepackage{color}
\usepackage{natbib}
\usepackage{fancyref}

\newcommand{\eqdef}{\stackrel{\mbox{\rm\tiny def}}{=}}
\usepackage{setspace}
\usepackage{amsthm}
\usepackage{abstract}

\theoremstyle{plain}
\newtheorem{assumption}{Assumption}
\newtheorem{theorem}{Theorem}

\renewcommand\appendix{\par
  \setcounter{section}{0}
  \setcounter{subsection}{0}
  \renewcommand\thesection{Appendix \Alph{section}}
  \renewcommand\thesubsection{\Alph{section}.\arabic{subsection}}
}

\usepackage{titlesec}
\titleformat{\section}
  {\normalfont\sffamily\Large\bfseries}
  {\thesection}{1em}{}

\titleformat{\subsection}
  {\normalfont\sffamily\large\bfseries}
  {\thesubsection}{1em}{}

\newcommand{\BE}{\begin{eqnarray}}
\newcommand{\EE}{\end{eqnarray}}
\newcommand{\mb}{\mathbf}
\newcommand{\argmin}{\operatornamewithlimits{argmin}}
\newcommand{\State}{\mathcal{S}}
\newcommand{\ACTION}{\mathcal{A}}
\newcommand{\mub}{\mu_{\mathrm{base}}}
\newcommand{\drp}{{\tt DRP}}
\newcommand{\rdrp}{{\tt RDRP}}
\newcommand{\ri}{{\tt RI}}

\begin{document}
%
\title{\sffamily \bfseries Optimal Demand Response Using Device Based Reinforcement Learning}
%
%
%

\author{Zheng~Wen,
        Daniel~O'Neill, 
        and~Hamid~Reza~Maei
\thanks{
Z. Wen, D. O'Neill and H. Maei 
are with the Department of Electrical Engineering, Stanford University,
Stanford, CA 94305 USA }
\thanks{
e-mail: zhengwen, dconeill, maei@stanford.edu
}
}

\date{}
\maketitle

\begin{abstract}
Demand response (DR) for residential and small commercial buildings is estimated to account for as much as 65\% of the total energy savings potential of DR,
and previous work shows that a fully automated Energy Management System (EMS) is a necessary prerequisite to DR in these areas.
In this paper, we propose a novel EMS formulation for DR problems in these sectors.
Specifically, we formulate a fully automated EMS's rescheduling problem as a reinforcement learning (RL) problem, and argue that this RL problem can be approximately solved by decomposing it over device clusters. 
%
%
Compared with existing formulations, our new formulation (1) does not require explicitly modeling the user's dissatisfaction on job rescheduling, (2) enables the EMS to self-initiate jobs,
(3) allows the user to initiate more flexible requests and (4) has a computational complexity linear in the number of devices.
%
%
We also demonstrate the simulation results of applying Q-learning,
one of the most popular and classical RL algorithms,
to a representative
example.
\end{abstract}


\maketitle

\section{Introduction}\label{introduction}
Demand response (DR) systems \cite{borenstein2002dynamic, braithwait2002role, 
barbose2004survey} dynamically adjust electrical demand in response to changing electrical 
energy prices or other grid signals. DR offers several benefits. By suitably adjusting energy 
prices, load can be shifted from peak energy consumption periods to other times. This, in 
turn, can  improve operational efficiency, reduce operating costs, improve capital 
efficiency, and reduce harmful emissions and risk of outages. The variability of renewables 
can create an additional need to shift energy consumption in order to better match energy 
demand with unforecasted changes in electrical energy generation. The benefit is a
reduction in backup (ancillary) generation frequently used to hedge renewable sources. 

There are several types of DR. In direct DR a utility or other entity directly modifies the energy consumption of users by adjusting the operation of user's equipment. Interruptible tariffs allow a utility to interrupt the supply of power to a company under predefined conditions.  Price driven DR uses pricing mechanisms to attempt to modulate energy demand. 
DR has been extensively investigated for larger energy users and has been implemented in many areas 
(\emph{e.g.},~\cite{651628,report}). Residential and small commercial building DR ~\cite{karen, Herter20072121,Faruqui200553,kochautomated} offers similar potential benefits.   DR for residential and small commercial buildings was estimated to account for as much as 65\% of the total energy savings potential of DR. However, DR in the residential and small commercial building sector faces several challenges. 

Technical challenges include deploying an infrastructure supplying real-time pricing information to energy consumers\footnote{Throughout this paper, we use the terms ``user" and ``consumer" interchangeably.} 
in a useful way, ensuring security, and implementing advanced metering and networking devices \cite{kochautomated, lemay2008}. In addition to all these technical challenges, another
challenge vital to the success of DR in the residential and small commercial building sectors is that it requires a fully automated Energy Management System (EMS) \cite{kochautomated,piettefield}. 
This is because
with price driven DR, consumers face a continuing sequence of 
decisions to either use a particular device now and consume energy at current (known)
prices or to defer using the device until later at possibly unknown prices. Each decision requires the consumer to weigh the cost differential against his dissatisfaction due to 
rescheduling device usage.  This is particularly burdensome when the consumer must also estimate future energy prices. Further many of these decisions have limited financial 
impact on the consumer \cite{oneill2010}, and, as a result, many rational
consumers in the residential and small commercial building sectors
may not be sufficiently incentivized to make these decisions over the long run (known as ``decision fatigue" in \cite{oneill2010}). 

To be effective, an EMS system needs to automatically make energy consumption decisions that are consistent with the cost delay trade-offs of energy users, in this way acting as an energy agent.    It is often difficult to cost-effectively model the behaviors of the idiosyncratic
consumers and the temporal variations of the energy prices, a successful DR EMS
needs to learn to make optimal decisions for consumers from interacting with the consumers and energy prices. Recently, O'Neill et al. \cite{oneill2010} proposed a fully-automated EMS algorithm based on reinforcement learning (RL), which 
learns how to make optimal decisions for consumers. To the best of our knowledge, this 
is the first paper to apply RL to DR in residential and small commercial building sector. The authors adopt a
request inventory model for the system dynamics and use Q-learning, a classical RL algorithm, to learn how to make the optimal decisions for energy consumers. In this approach, users make energy requests to the EMS system (e.g. pushing a button on a device that a user wants to run) and the system schedules the time of operation by calculating the user's trade-offs between delay and energy prices.  It learns these trade-offs by observing energy consumer's behaviors and observing the patterns of energy pricing. Over time the EMS learns to make the best decisions for energy users in the sense that it balances energy cost and the delay in energy usage in the same way that the customer optimally would, but without the consumer having to make the decision. The authors explicitly assume that (1) consumers' dissatisfaction with delay can be modeled by known disutility functions, and (2) that consumers explicitly initiate all energy usage.
This approach has several limitations:
\begin{itemize}
\item Finding specific disutility functions for a particular residence or small business can be difficult and costly. In  \cite{oneill2010}, the authors assume these functions have particular mathematical properties, but do not address how these functions might be determined.  These functions are likely to be idiosyncratic and are specific to energy price vs. time delay trade-offs. 
\item Many energy consuming activities occur without the consumer directly initiating them. HVAC in office buildings and pool heaters in residential settings are obvious examples. A useful EMS would self-initiate jobs for these and similar devices without explicit user requests or reservations.  For example, if it is unexpectedly hot in a summer afternoon and the current energy price is expected to be cheaper than that in the evening, then the EMS should be allowed to turn on AC without an explicit request/reservation from the consumer. 
\item  The computational complexity of this approach grows exponentially as the number of devices. Known as the ``curse-of-dimensionality" in dynamic programming (DP) and RL literature, 
this problem limits the approach to fairly small numbers of devices.
\end{itemize}

In this paper, we propose a novel EMS formulation that addresses the limitations of  \cite{oneill2010} described above.
Our proposed algorithm also uses RL, but adopts a device centric point of view, and, as we will discuss in detail in Section \ref{sec:general}, under reasonable assumptions, the RL problem decomposes over \emph{device clusters} and it is sufficient to apply an RL algorithm to each individual device cluster.
Specifically, the algorithm addresses these issues:
\begin{enumerate}
\item Our EMS formulation does not require a pre-specified disutility function modeling the consumer's dissatisfaction on job rescheduling. Instead, under this formulation,
the EMS is able to learn the consumer's dissatisfaction 
based on
his evaluations on completed/canceled jobs. In other words, our new RL formulation has eliminated the impractical assumption in \cite{oneill2010} that consumers' dissatisfactions with delay can be captured by known disutility functions.
\item Our approach allows both user-initiated jobs and EMS-initiated jobs.  The EMS-initiated jobs use a probing/feedback mechanism to find the best way to anticipate future energy usage. 
\item Our EMS algorithm also enables more flexible user-initiated jobs, specifically
\begin{itemize}
\item A consumer request's \textit{target time} can be different from its \textit{request time}, where the request time is the time
when the EMS receives this consumer request, and the target time is the time when the consumer prefers
this request to be satisfied.
\item Consumer requests/reservations can have different priorities, whereas in \cite{oneill2010}, all the consumer requests have the same priority.
\item Energy requests can be canceled by the consumer, reflecting the behavior of real energy users.
\end{itemize}
\item  The computational complexity of our approach grows linearly with the number of device clusters,
and thus many classical RL algorithms can be applied even when there are a large number of devices.
\end{enumerate}

In this paper, we also propose new performance metrics for
RL algorithms applied to this problem. In particular, we suggest methods of measuring performance relative to the user's current pattern of behavior and relative to a prescient optimal pattern of behavior. 

Before proceeding, we briefly review some relevant literature.
\cite{kara2012} also applies RL techniques to a smart grid application. In particular, it focuses on how to use RL to control a population of heterogeneous thermostatically-controlled loads. Another directly relevant paper is \cite{Turitsyn2011}, which also proposed device based Markov decision process (MDP) models. Compared with \cite{Turitsyn2011}, this paper is new in the following three points: first, this paper motivates and discusses why the optimal DR problem is approximately decomposed over device clusters, while \cite{Turitsyn2011} does not include such motivation/discussion and directly focuses on device based MDP models without justifying why this approach is reasonable.
Second, \cite{Turitsyn2011} assumes that the models of the user behavior and grid signals are known. As we have discussed above, such assumptions are not realistic in practice. In this paper, we use RL techniques to learn such models. Finally, though the MDP model proposed in this paper is still a simplified model, it is much more general and realistic than the models proposed in \cite{Turitsyn2011}.

The remainder of this paper is organized as follows. 
In Section \ref{ems_overview}, 
we briefly describe how a practical fully-automated EMS
should interact with the consumer and the grid signals.
In Section \ref{sec:general}, we argue that the optimal demand response problem is approximately decomposed over
device clusters, and pose it as a collection of device based RL problems.
In Section \ref{sec:specific}, we motivate and propose a simplified MDP model for a device based RL problem, and discuss how to extend it to more 
general models.
Then, in Section \ref{sec:experiment}, we demonstrate the simulation results on a representative example when the classical Q-learning 
is applied.
We conclude and discuss future work in Section \ref{sec:conclusion}.

\section{Description of fully-automated EMS}\label{ems_overview}

A fully-automated EMS (henceforth referred to as EMS) is a necessary prerequisite to DR
in the residential and small commercial building sectors. Furthermore, an EMS needs to learn how to make the optimal decisions for 
the user while interacting with them and the real-time grid signals.
In this section, we describe how a fully-automated EMS should interact with the user and the grid signals.

\begin{figure}[h]
\centering
\includegraphics[scale=0.5]{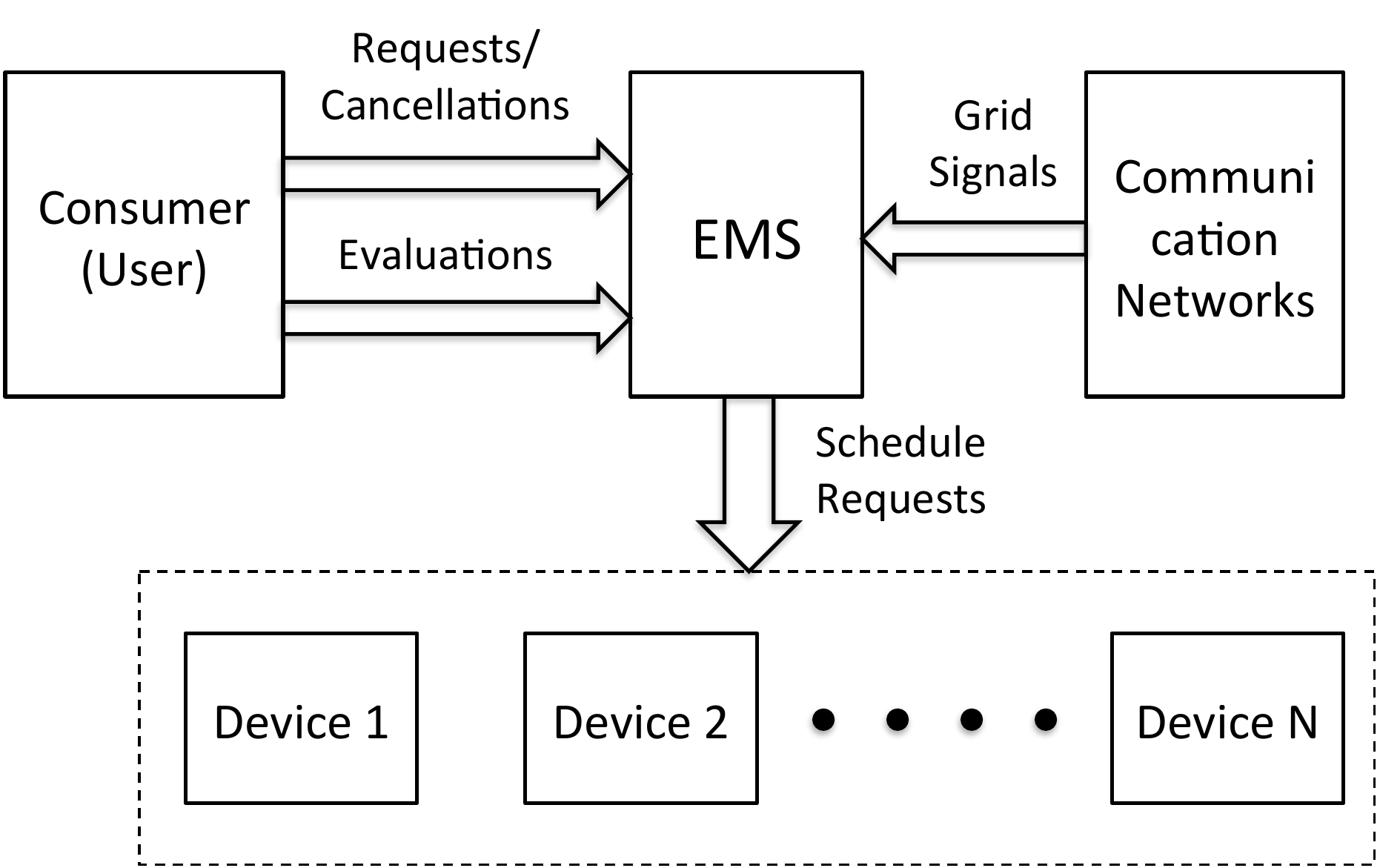}
\caption{Fully-Automated EMS}
\label{f1}
\end{figure}

Generally speaking, a fully-automated EMS observes the grid signals, receives requests and evaluations from the user, and schedules the jobs over the devices (Figure \ref{f1}). We detail the interaction mechanisms in the remainder of this subsection.

\subsubsection*{Grid Signals}
The EMS observes the grid signals through a communication network, where the ``communication network" refers to the infrastructure that supplies  grid signals to the EMS. Any exogenous information that is effectively delivered by the communication network and is
useful for the EMS to make the scheduling decisions can be regarded as a grid signal.
The most common grid signal is the real-time energy price; other grid signals might include the expected future energy prices, the real-time temperature and weather condition, and other useful exogenous information.

Notice that most grid signals are \emph{exogenous} in the sense that the EMS's actions will not influence them. In particular, we assume that the energy price is exogenous (i.e. the EMS 
is a \textit{price-taker}). This assumption is reasonable since in the residential and small commercial building sectors, the \textit{market power} of an individual EMS (or equivalently, of an individual consumer) is so small that the impact of its actions on energy prices is negligible.

\subsubsection*{User-Initiated Jobs and EMS-Initiated Jobs}
A fully-automated EMS should perform the following 
functions:
\begin{itemize}
\item The EMS receives requests from the consumers, and then schedules when to fulfill the received requests. We henceforth refer to this case as
 a \textit{user-initiated job}. We further assume that the EMS allows a customer to cancel existing uncompleted requests.
\item If a smart device managed by the EMS
is idle (i.e. currently there is no request for that device), the EMS could speculatively power that device.
We henceforth refer to this case as an \textit{EMS-initiated job}. For instance, in a small commercial building, the EMS might speculatively turn on the building's air conditioning in advance of the tenant's arrival to capture early morning lower energy costs or to mask the latency of cooling the building. Notice that we should not
allow the EMS to do speculative jobs on all the smart devices (such as dishwashers).
\end{itemize}
We assume time is discrete $t=0,1,\cdots$ and that there are $N$ smart devices
managed by the EMS and numbered $n=1,2,\cdots, N$. 
To simplify exposition, we assume that all the jobs done by device $n$ are standardized and hence
they can be completed in one time step and consume a constant energy $C_n$, which only depends on the type of the smart device. 
This assumption can be readily relaxed to devices with different operating periods.

\subsubsection*{Interaction between user and EMS}
We now describe how the EMS interacts with the consumer (user) in a user-initiated job and an EMS-initiated job. 
Specifically, as its name suggests, a user-initiated job starts with a user sending a request to the EMS.
Specifically, 
each consumer request 
is represented by a four-tuple $J=(n, \tau_r, \tau_g, g )$, where 
\begin{itemize}
\item $n$ 
denotes the requested device;
\item $\tau_r$ is the \textit{request time} and denotes when the EMS receives this request;
\item $\tau_g$ is the \textit{target time} and denotes when the user prefers this requested job to be completed;
\item $g$ denotes the \textit{priority} of this request, with higher priority implying the ``stronger preference" of the
user that they want the requested job to be completed at a time close to the target time $\tau_g$.
\end{itemize}
Notice that the target time $\tau_g$ is not necessarily equal to the request time $\tau_r$; instead, the user might request to use a device in a later 
time (i.e. $\tau_g \ge \tau_r$). On the other hand, for a request 
$J=(n, \tau_r, \tau_g, g )$, it is unreasonable to assume that $\tau_g-\tau_r$, the difference between the target time and the request time, can
be arbitrarily large. Thus, in this paper, we assume that
(1) for any request $J=(n, \tau_r, \tau_g, g)$, its target time $\tau_g$ must satisfy
$\tau_r \le \tau_g \le \tau_r +W_n$, and (2) if the request $J=(n, \tau_r, \tau_g, g)$
is not fulfilled by time $\tau_g +W_n$, then it will be automatically canceled by the EMS, where $W_n$ is a known time window and only depends on the type of
device.

We also assume that an unsatisfied consumer request can be canceled by the user.
Furthermore, we assume that at some time (e.g., at the end of a day), the user will evaluate some (not necessarily all) completed/canceled requests. As we will see later, the EMS can use such evaluations to learn the user's dissatisfaction on the rescheduling of the user-initiated jobs.

On the other hand, an EMS-initiated job is started by the EMS, without receiving a request from the user. The only interaction between the EMS and the user for such jobs is that the user will evaluate some EMS-initiated jobs at some time.
As is in the user-initiated jobs, the EMS also exploits such evaluations to learn user's dissatisfaction with the EMS-initiated jobs.

In summary, the interaction between user and EMS is as follows: for user-initiated jobs, the possible interactions include
(1) the user sends requests to the EMS, (2) the user can choose to cancel the unsatisfied requests and (3) the user evaluates 
some completed/canceled requests. On the other hand, for EMS-initiated jobs, the only interaction between user and EMS is that
the user evaluates some completed EMS-initiated jobs. 

\section{Device Based Reinforcement Learning}
\label{sec:general}
This section proceeds as follows. We first motivate and define the dissatisfaction function and instantaneous cost function in Subsection \ref{sec:dissatisfaction}. Explicit dissatisfaction functions are not required in practice, but
we assume they are known in the first two subsections to facilitate easy exposition of the problem.
Then, Subsection \ref{sec:decomposition} decomposes the optimal demand response problem into a collection of device based MDPs under suitable assumptions. We motivate and propose 
the reinforcement learning formulation for a device based MDP
in Subsection \ref{sec:rl}.

\subsection{Dissatisfaction Function and Cost Function}
\label{sec:dissatisfaction}
To formalize the notion of optimal demand response, in this subsection,
we define the dissatisfaction function for a consumer (user) that captures their
preferences (dissatisfaction) over job rescheduling.

Let $\mathcal{H}_t$ denote the ``history" of device operations, consumer requests/cancellations and the EMS decisions by the end of time period $t$. 
Then the user's dissatisfaction on rescheduling at time $t$ should 
be a function of 
$\mathcal{H}_t$, and we denote this function as
$\bar{U}^{(t)} \left  ( \mathcal{H}_t \right )$.
Furthermore, we use $\mathcal{H}_t^{(n)}$ to denote the ``history"
of device operations, consumer requests/cancellations and the EMS decisions
for device $n$ by the end of time period $t$.

Obviously, directly working with $\bar{U}^{(t)} \left  ( \mathcal{H}_t \right )$  will result in a computationally intractable problem. To overcome this challenge, in this paper, in this paper, we make the 
following simplifying assumption:
\begin{assumption}
\label{assump:assump1}
For any $t\ge 0$, the dissatisfaction function $ \bar{U}^{(t)}$ is approximately additive over the devices, that is
\BE
\bar{U}^{(t)} \left ( \mathcal{H}_t  \right) \approx \sum_{n=1}^N \bar{U}^{(t,n)} (  \mathcal{H}_t^{(n)} ),
\EE
where $\bar{U}^{(t,n)}$ captures the consumer's dissatisfaction at time $t$ for device $n$ and $\mathcal{H}_t^{(n)}$ is the ``history" for device $n$ by time
$t$.
\end{assumption}
Assumption \ref{assump:assump1} is motivated by the observation that a \emph{rational} consumer's preference over job rescheduling is weak compared to his preferences in other aspects of life\footnote{
Let us provide an intuitive motivation for Assumption \ref{assump:assump1} from the following perspective:
let $\bar{\mb{u}}_{\mathrm{DR}} \in \Re^N$ be a vector encoding the user's dissatisfactions on job rescheduling over $N$ devices, and $\bar{\mb{u}}_{\mathrm{other}}$  be a vector encoding the user's dissatisfactions in other aspects of life. Assume the overall dissatisfaction (unhappiness) of the user is $f(\bar{\mb{u}}_{\mathrm{DR}}, \bar{\mb{u}}_{\mathrm{other}})$, where $f$ is a general non-linear function. Now we consider $f(\bar{\mb{u}}_{\mathrm{DR}}+\Delta \bar{\mb{u}}_{\mathrm{DR}}, \bar{\mb{u}}_{\mathrm{other}})$, notice that the weak preference over job rescheduling implies that $\Delta \bar{\mb{u}}_{\mathrm{DR}}$
is ``small", thus, $f(\bar{\mb{u}}_{\mathrm{DR}}+\Delta \bar{\mb{u}}_{\mathrm{DR}}, \bar{\mb{u}}_{\mathrm{other}}) $
can be well approximated by
$
f(\bar{\mb{u}}_{\mathrm{DR}}, \bar{\mb{u}}_{\mathrm{other}})+ 
\nabla_{\bar{\mb{u}}_{\mathrm{DR}}} f(\bar{\mb{u}}_{\mathrm{DR}}, \bar{\mb{u}}_{\mathrm{other}})
\Delta \bar{\mb{u}}_{\mathrm{DR}}
$, where $\nabla_{\bar{\mb{u}}_{\mathrm{DR}}} f$ is the partial derivative vector of $f$ with respect to $\bar{\mb{u}}_{\mathrm{DR}}$.  Notice that for our purpose, we only care about the ``change" of the user's overall unhappiness as a function of the ``changes"
in his dissatisfactions on job rescheduling, thus, the user's dissatisfaction function is
$\bar{U}(\Delta \bar{\mb{u}}_{\mathrm{DR}})=f(\bar{\mb{u}}_{\mathrm{DR}}+\Delta \bar{\mb{u}}_{\mathrm{DR}}, \bar{\mb{u}}_{\mathrm{other}})-f(\bar{\mb{u}}_{\mathrm{DR}}, \bar{\mb{u}}_{\mathrm{other}}) \approx \nabla_{\bar{\mb{u}}_{\mathrm{DR}}} f(\bar{\mb{u}}_{\mathrm{DR}}, \bar{\mb{u}}_{\mathrm{other}})
\Delta \bar{\mb{u}}_{\mathrm{DR}}$, which is a weighted sum of $\Delta \bar{\mb{u}}_{\mathrm{DR}}$.
In Assumption \ref{assump:assump1}, we further simplify the dissatisfaction function by assuming that all the weights are equal to $1$.
} (see the justification on ``decision fatigue" in \cite{oneill2010}).

With the dissatisfaction function $\bar{U}^{(t)}$ defined above,
we further assume the instantaneous cost function of the consumer at time $t$ has the following form:
\BE
P_t \sum_{n \in \mathcal{D}(t) }  C_n+\gamma \bar{U}^{(t)} \left  ( \mathcal{H}_t  \right),
\label{disutility_1}
\EE
where $\mathcal{D}(t) = \left \lbrace  \textrm{devices that do a job at time $t$}\right \rbrace$,
$P_t$ is the electricity price at time $t$, and 
$\gamma>0$ represents the tradeoff between the electricity bill paid and the 
consumer's dissatisfaction on rescheduling.
Specifically,
notice that $\sum_{n \in \mathcal{D}(t) }  C_n$ is the total electricity energy consumed at time $t$, and hence $P_t \sum_{n \in \mathcal{D}(t) }  C_n$
is the electricity bill paid at time $t$. 

Note that from the EMS's perspective, both the electricity price and the consumer behavior are exogenous and stochastic; thus, in this paper, we 
assume that EMS aims to minimize the expected infinite-horizon discounted cost:
\BE
\mathbb{E} \left \{ \sum_{t=0}^{\infty} \alpha^t \left[  P_t \sum_{n \in \mathcal{D}(t) }  C_n+\gamma \bar{U}^{(t)} ( \mathcal{H}_t )  \right] \right \}, \nonumber
\EE
where $0<\alpha<1$ is a discrete-time discount factor.
Under Assumption \ref{assump:assump1}, the infinite-horizon
discounted cost function can be approximated by
\BE
\sum_{n=1}^N \left[
\mathbb{E} \left \lbrace
\sum_{t=0}^{\infty} \alpha^t \left[ P_t C_n \mathbf{1}(n \in \mathcal{D}(t))
+ \gamma 
\bar{U}^{(t,n)} \left  ( \mathcal{H}_t^{(n)} \right) \right] \right \rbrace \right], \label{disutility_4B}
\EE 
which decomposes over devices.

\subsection{Decomposition of the Problem}
\label{sec:decomposition}
In this subsection, we first propose a dynamic model for the grid signals.
Then, we motivate an assumption on user behavior (Assumption \ref{assump:assump3}), and show that under the proposed assumptions, the
optimal demand response problem (i.e. the EMS's job scheduling problem)
is approximately decomposed over device clusters. 

Let us start by proposing a dynamic model for the grid signals (e.g. energy price, temperature).
As discussed in Section \ref{ems_overview}, grid signals are exogenous in the
sense that the EMS's actions will not influence them. 
As is discussed in some recent literature (e.g. \cite{oneill2010}),
grid signals can be modeled as Markov processes.
Thus, throughout this paper, we make the
following assumption:
\begin{assumption}
\label{assump:asump2}
All the grid signals follow exogenous Markov processes.
\end{assumption}
Notice that in general, these Markov processes are time-dependent.
We now propose an assumption on user behavior under which the EMS's 
job scheduling problem is approximately decomposed over device clusters.
\begin{assumption}
\label{assump:assump3}
The $N$ devices can be classified into clusters such that conditioning on time and grid signals, 
the consumer requests/cancellations to different device clusters are weakly statistically dependent.
\end{assumption}

Assumption \ref{assump:assump3} is motivated by daily observations.
For instance, 
in a residential household, we can classify the air-conditioner, 
electric vehicle, laundry machine and dryer into three clusters: cluster 1 includes the air-conditioner; cluster 2 includes the laundry machine and dryer; and
cluster 3 includes the electric vehicle. Under this clustering,
it is observed that for most residential consumers, conditioning on time and grid signals (especially the energy price), the statistical dependence between 
their requests to devices in different clusters are weak.
 
Recall that the EMS's objective is to find an optimal scheduling policy to minimize (\ref{disutility_4B}), which approximately decomposes over devices under Assumption \ref{assump:assump1}. 
Assumption \ref{assump:assump3} states that the consumer requests/cancellations to different device clusters are almost conditionally independent. Thus, under
Assumptions \ref{assump:assump1}-\ref{assump:assump3}, the EMS's job scheduling problem is approximately decomposed over device clusters.
Specifically, 
for each device cluster $\mathcal{C} \subseteq \left \{1,2,\ldots, N \right \}$,
its job scheduling problem is characterized by
\begin{itemize}
\item discount factor $\alpha$, tradeoff parameter $\gamma$;
\item dissatisfaction function $\bar{U}^{(t,n)}$ and energy consumption rate $C_n$ for each device $n$ in the cluster; 
\item exogenous Markov processes modeling grid signals;
\item statistical model of user requests/cancellations to devices in this cluster.
\end{itemize}
The objective of this job scheduling problem is to minimize
\BE
\sum_{n \in \mathcal{C}} \left[
\mathbb{E} \left \lbrace
\sum_{t=0}^{\infty} \alpha^t \left[ P_t C_n \mathbf{1}(n \in \mathcal{D}(t))
+ \gamma 
\bar{U}^{(t,n)} \left  ( \mathcal{H}_t^{(n)} \right) \right] \right \rbrace \right], \label{disutility_4C}
\EE 
As we will discuss in Section \ref{sec:specific}, by properly specifying the statistical model of user requests/cancellations, this job scheduling problem can be modeled as an MDP.
We henceforth refer to this MDP as the device based MDP and focus on it in the remainder of this paper.

Since the optimal demand response problem is approximately decomposed over device clusters, thus, a near-optimal solution can be obtained by solving all the decomposed device based MDPs separately. Obviously, the computational complexity to derive this near-optimal solution is linear in the number of device clusters.

\subsection{Reinforcement Learning Formulation}
\label{sec:rl}

\begin{figure}[h]
\centering
\includegraphics[scale=0.75]{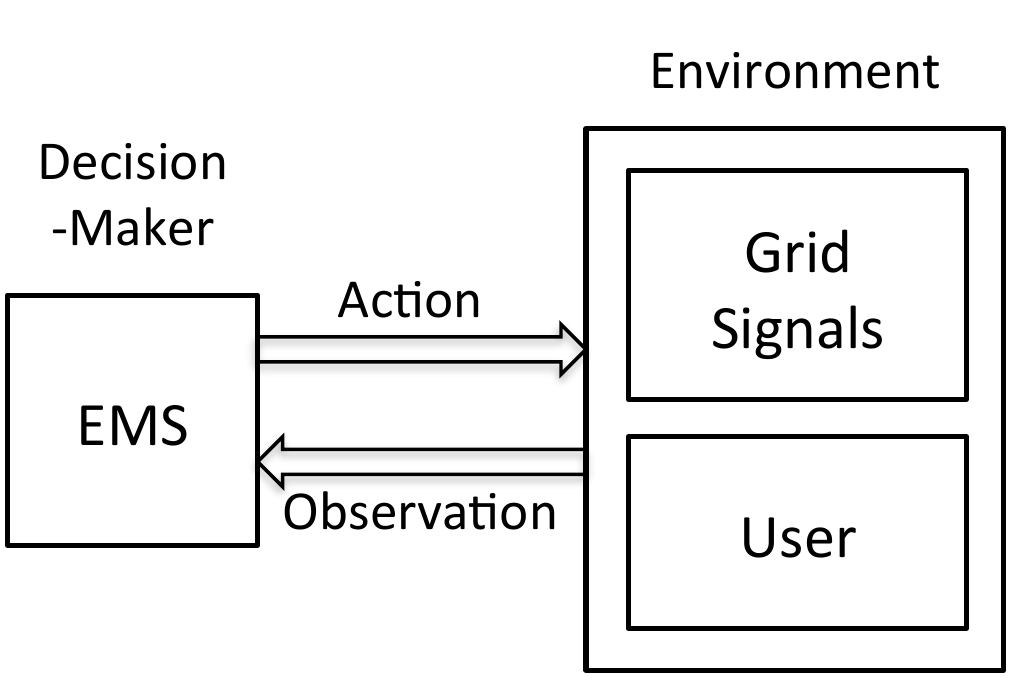}
\caption{Device Based Reinforcement Learning
}
\label{f5}
\end{figure}

However, as we discussed in Section \ref{introduction}, in most practical cases,
the dissatisfaction function of the user and the statistical models of the grid signals and user requests/cancellations are initially
unknown. Thus, the EMS must learn how to make optimal decisions for a device cluster based on the incrementally gathered data from the user and
the grid signals. Reinforcement learning (RL) is a collection of techniques for a decision-maker to learn
how to make optimal decisions while interacting with an unknown ``environment" 
\cite{sutton1998reinforcement,Bertsekas1996,2010Szepesvari,wen2014dissertation,WenVR13_deterministic}. RL tries to strike a balance between learning (exploration) 
and optimization (exploitation), and has been extensively used in many other fields, 
such as artificial intelligence and robotics \cite{CoatesAN10} and petroleum engineering \cite{WenDVA2011,WenDVA2012}. 
As we have discussed in Section \ref{introduction},
some recent literature has also applied RL to some 
smart grid applications \cite{kara2012}.
As has been discussed in \cite{oneill2010}, it is natural to use RL to solve an EMS's learning problem
since EMS needs to learn to make decisions while interacting with the user and the grid signals. As is shown in Figure~\ref{f5}, under the RL formulation of the optimal DR problem, the agent is the EMS, and the environment includes both the grid signals and the user.
Since this RL problem focuses on a device cluster, we refer to it as the device based RL problem.

In practice, the statistical models of the grid signals and user requests/cancellations can be learned by directly observing the user behavior and the grid signals. As we will discuss in Section \ref{sec:specific}, under reasonable assumptions, the user's dissatisfaction function can be learned based on the user's evaluations on completed/canceled jobs.

It is worth pointing out that any RL algorithm (see \cite{sutton1998reinforcement,Bertsekas1996,2010Szepesvari,wen2014dissertation}) can be applied to the device based RL problem. To demonstrate the performance of an RL algorithm, we need to compare it with a reasonable \emph{baseline policy}.
Since this paper focuses on how demand response (DR) can potentially reduce a user's (expected infinite-horizon discounted) cost, we choose
the baseline policy as the one without job rescheduling. Specifically,
under this baseline policy, the EMS never self-initiates a job, and all the jobs initiated by the user are scheduled to be completed at their target times. We use this baseline policy in Section \ref{sec:experiment}.

\section{A Simplified MDP Model}
\label{sec:specific}
This section proceeds as follows. We first propose a simplified device based MDP model in Subsection \ref{sec:simplifying} to \ref{sec:mdp_model}. Then, in Subsection \ref{sec:drp} and \ref{sec:metrics}, we motivate and define a performance metric for RL algorithms applied to the proposed device based MDP model. Finally, we discuss how to extend the proposed MDP model to more general models in Subsection \ref{sec:extension}.

It is worth pointing out that all the assumptions in this section are made to simplify the exposition of the device based MDP model. These simplifying assumptions are nonessential in the sense that as long as the assumptions proposed in the previous section hold, the EMS's 
job scheduling problem is approximately decomposed over device clusters. 
We will discuss how to relax some of these simplifying assumptions in Subsection \ref{sec:extension}.

\subsection{Simplifying Assumptions}
\label{sec:simplifying}
We start by making some simplifying assumptions. 
First, we assume that each device cluster has only one device. Thus, we focus on deriving an optimal scheduling policy for a single device in this section.
Second, we assume that the only available grid signal is the real-time energy price, and the
statistical models of the energy price and user behavior\footnote{By ``statistical model of user behavior", we mean statistical model of user requests/cancellations/evaluations to that device.} are time-invariant.
Third, we also assume that the user will not submit a new request to a device if that device currently has an uncompleted request.\footnote{Notice that this assumption does not rule out \emph{request replacement}. Specifically, the user can replace an uncompleted request with a new request by first canceling the uncompleted request and then submitting the new request.}
Finally, we assume that the user will evaluate all the completed/canceled jobs immediately after the job is completed/canceled. 
As we will detail in Subsection \ref{sec:extension}, all these simplifying assumptions can be effectively relaxed.

\subsection{Functional Form of Dissatisfaction Function}
\label{sec:functional_form}
In this subsection,
we propose a specific functional form for the dissatisfaction function $\bar{U}^{(t,n)}$, for any device $n$.
Since a rational consumer's preference over job rescheduling is weak compared to his preferences in other aspects of life,
we assume that 
\begin{itemize}
\item If there is no job completed/cancelled at time $t$ on device $n$, we assume that $\bar{U}^{(t,n)}  (  \mathcal{H}_t^{(n)}  )=0$. 
\item If there is a job completed/cancelled at time $t$ on device $n$, we assume that
$\bar{U}^{(t,n)}  (  \mathcal{H}_t^{(n)} )$
only depends on the job completed/canceled. That is, $\bar{U}^{(t,n)}  (  \cdot )$ is time-invariant and does not depend on the 
previously completed/canceled jobs. Specifically, if device $n$ satisfies consumer request $J(t,n)=(n, \tau_r, \tau_g, g )$ at time $t$, we assume 
\BE
\bar{U}^{(t,n)}  (  \mathcal{H}_t^{(n)} ) =\tilde{U}^{(n)}_r(t-\tau_g, g),  \label{U_r}
\EE
where the subscript ``$r$" denotes that it is the dissatisfaction incurred when a request is satisfied, and $g$ is the priority
of the request. Note $t-\tau_g$ captures 
not only the distance between the current time $t$ and the request's target time $\tau_g$, but also whether or not the target time
has passed. Similarly, if the request $J(t,n)=(n, \tau_r, \tau_g, g )$ is canceled by the consumer at time $t$, we assume
\BE
\bar{U}^{(t,n)}  (  \mathcal{H}_t^{(n)} ) =\tilde{U}^{(n)}_c(t-\tau_g, g), \label{U_c}
\EE
where the subscript ``$c$" denotes that it is the dissatisfaction incurred when a request is canceled.
Finally,
if EMS initiates a job on device $n$ at time $t$, we assume that
\BE
\bar{U}^{(t,n)}  (  \mathcal{H}_t^{(n)} )  =\tilde{U}^{(n)}_e (t-\tau_p), \label{U_s}
\EE
where the subscript ``$e$" denotes that it is the dissatisfaction incurred when an EMS-initiated job is done, and $\tau_p-1$ is the time
when the previous job on device $n$ (either user-initiated or EMS-initiated) is completed or cancelled. 
\end{itemize}
Note that the dissatisfaction function class described in Equations (\ref{U_r}-\ref{U_s}) is still quite general, and it
is very challenging to proceed to derive the specific functional
forms of $\tilde{U}^{(n)}_r$, $\tilde{U}^{(n)}_c$ and $\tilde{U}^{(n)}_e$.
In practice, $\tilde{U}^{(n)}_r$, $\tilde{U}^{(n)}_c$ and $\tilde{U}^{(n)}_e$ should be learned based on the user's evaluations on completed/canceled jobs. In this section, we make the following assumption:
\begin{assumption}
\label{assump:assump4}
The user's dissatisfaction on the rescheduling of a completed/canceled job
is equal to their evaluation on this job.\footnote{Assumption \ref{assump:assump4} is an idealized assumption on the relationship between
evaluation and dissatisfaction.
In practice, it might be more reasonable to assume that 
the user's evaluation on a job is equal to their dissatisfaction on the rescheduling of this job plus a (zero-mean, finite-variance, statistically independent)
``behavioral noise". However, the ``behavioral noise" does not fundamentally change the assumption
and we make Assumption \ref{assump:assump4} to simplify exposition.}
\end{assumption}
We now briefly motivate Assumption \ref{assump:assump4}: for a rational user, their evaluation on a completed/canceled job mainly reflects their dissatisfaction on this job,
and this dissatisfaction is either due to the (high) energy price or the (undesirable) rescheduling of the job. 
As we have discussed in Section
\ref{introduction}, due to ``decision fatigue",
most rational users will not reference the current energy price or any expected future energy prices before they send their evaluations to the EMS. Thus, their evaluations will
mainly reflect their dissatisfactions on job rescheduling. Thus, it is reasonable to assume Assumption \ref{assump:assump4}.

\subsection{MDP Model}
\label{sec:mdp_model}
We now propose a simplified MDP model for the device based RL problem, under the simplifying assumptions proposed in Subsection \ref{sec:simplifying}. We also assume that the user's dissatisfaction function has the functional form specified by (\ref{U_r}-\ref{U_s}). Thus, we only need to specify the statistical models of the energy price and user requests/cancellations. Since we focus on deriving an optimal scheduling policy for
a single device, we drop the subscript $n$ to simplify the exposition. For example, we will use 
$W$ instead of $W_n$ to denote the time window and represent a request as $J=(\tau_r, \tau_g, g)$.
We also use the 
term ``smart device" and ``EMS" interchangeably in this section, since one can think each
smart device (i.e. each device cluster in this case) has its own EMS.

First, we assume that the energy price $P_t$ follows an exogenous finite-state ergodic Markov chain. We use $\mathcal{P}$ to denote the set of states of this Markov chain, and use $|\mathcal{P}|$ to denote the 
number of states.

We model the user requests/cancellations to the smart device as a controlled stochastic process.
Moreover, we make the following simplifying assumption on our proposed MDP model:
\begin{assumption}
\label{assump:assump5}
The consumer requests/cancellations are statistically independent of the energy price process
$P_t$.
\end{assumption}
We now briefly motivate Assumption \ref{assump:assump5}. Notice that on one hand, as we have discussed in Section  \ref{ems_overview}, in the residential and small building sector, the market power of an individual
consumer is so small that the impact of their behavior on $P_t$ is negligible. On the other hand, as we have discussed in Section \ref{introduction}, due to ``decision fatigue", most rational consumers are not incentivized to reference  energy prices
before sending requests/cancellations to the EMS. In other words, for such consumers, their requests/cancellations are 
not influenced by energy prices. Thus, it is reasonable to assume Assumption \ref{assump:assump5}.

\begin{figure}[h]
\centering
\includegraphics[scale=0.5]{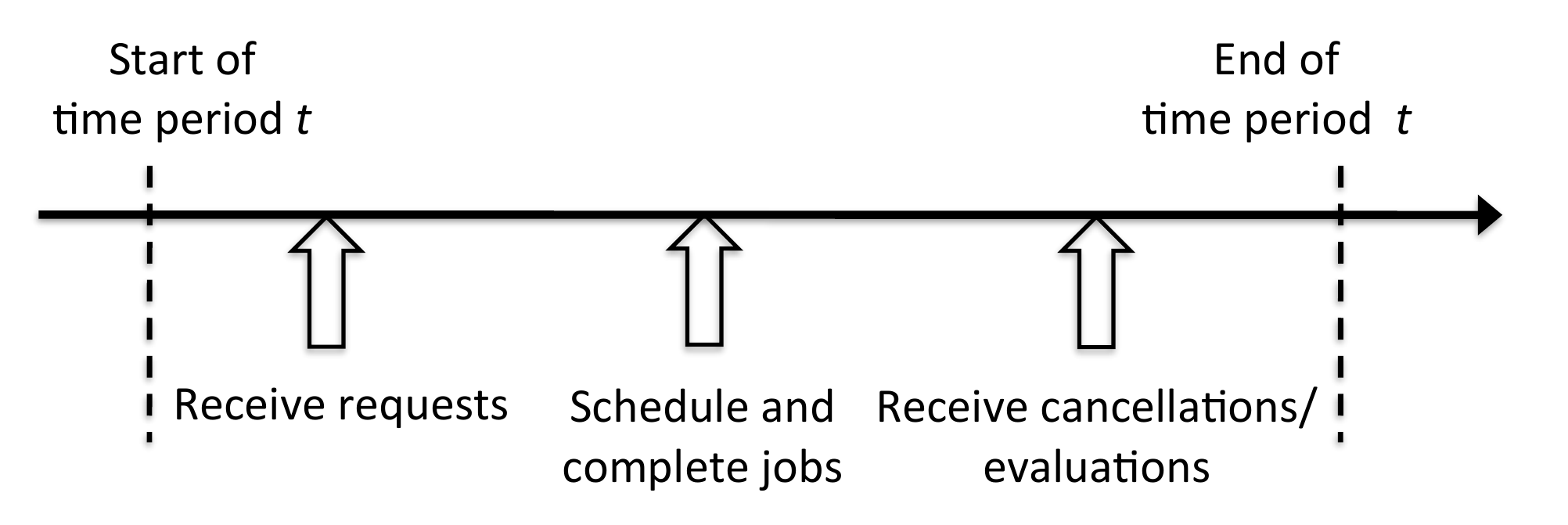}
\caption{User-EMS interaction timeline in time period $t$}
\label{f2}
\end{figure}

\begin{figure}[h]
\centering
\includegraphics[scale=0.7]{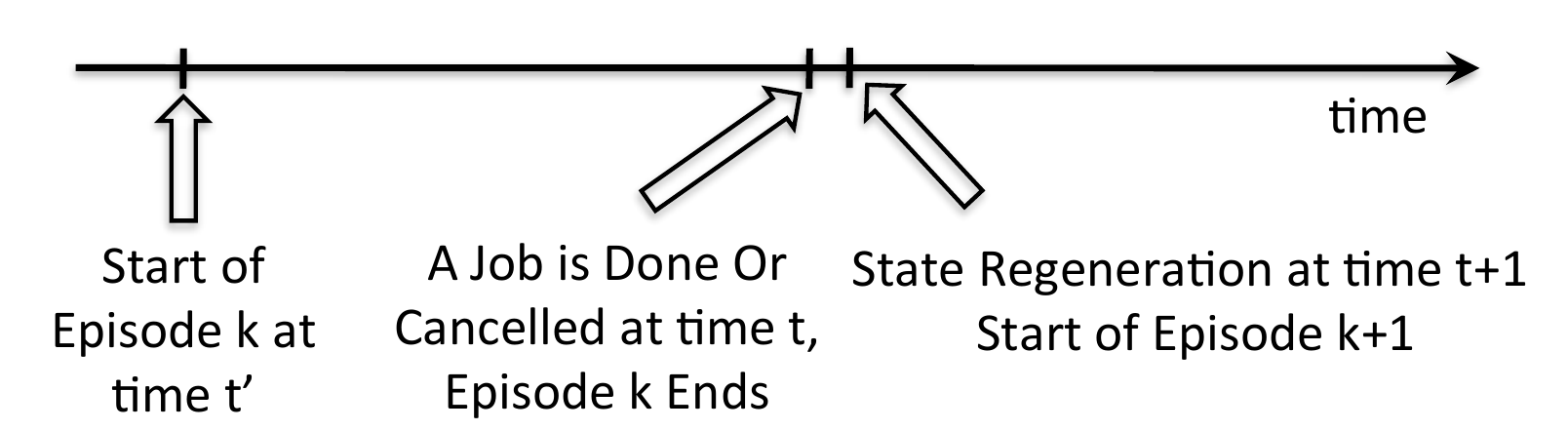}
\caption{Illustration of the notion of episode, where episode $k$ starts at time $t'$ and ends in time $t$
}
\label{f4}
\end{figure}

The ``user-EMS interaction timeline" in a single time
period\footnote{That is, whether the EMS receives the user requests, cancellations, and evaluations before or after it schedules jobs in that time period.} is illustrated in Figure \ref{f2}.
We further assume that the timeline for a smart device can be divided into ``episodes", which in general consists of multiple time steps.
Specifically, we assume that whenever the smart device completes a job (either initiated by the user or by the EMS) or the current 
unsatisfied request is canceled by the user, the 
current episode terminates. In the next time step, the smart device ``regenerates" its state according to a fixed 
distribution $\pi_0$ and a new episode starts.
The notion of episode is illustrated in Figure \ref{f4}.

We now propose an
 MDP model for a single smart device.
The state of the MDP at time $t$ is
$ x_t=\left[
  P_t, s_t, g_t
\right]^T \in \State$, 
where $P_t$ is the exogenous electricity price at time $t$, $s_t$ is the \textit{elapsed time} at time $t$, 
$g_t$ is the priority of request at time $t$ and $\State$ is the state space. Specifically, we define
\BE
s_t=\left \lbrace
\begin{array}{ll}
t-\tau_p & \textrm{if no request received in the current episode} \\
t-\tau_g  & \textrm{otherwise}
\end{array}
\right . \label{eq:elapsed_time}
\EE
where $\tau_p$ is the start time of the current episode and $\tau_g$ is the target time of the received request.
Furthermore, we use $g_t=0$ to denote that no request has yet been received in the current episode;
once a request is received in the current episode,
 we assume its priority $g_t \in \lbrace 1, 2,\cdots, g_{max} \rbrace$.

\begin{figure}[h]
\centering
\includegraphics[scale=0.5]{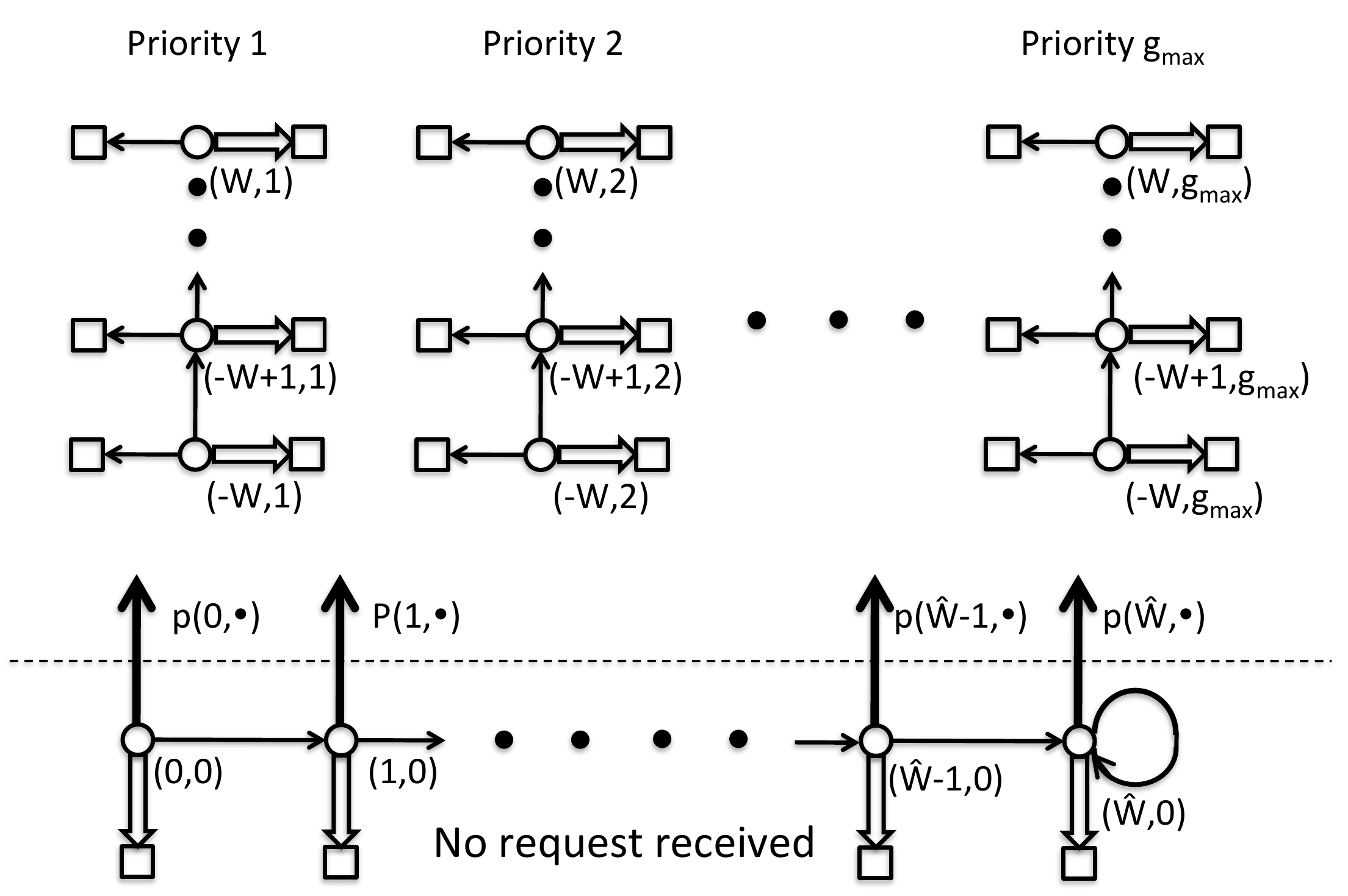}
\caption{The state transition model of the ``device portion". Notice that each circle corresponds to a ``device portion" state 
$\left[ s_t, g_t \right]^T$ and each square corresponds to the termination of the current episode and regeneration in the next time step.
The hollow arrows indicate the state transitions under action ``on", while the line arrows indicate the state transitions under action ``off".
The bold line arrows across the dotted line
indicate the fact that the states below the dotted line can transit to many states above the dotted line, since there are $(W+1)g_{max}$ types of requests.
}
\label{f3}
\end{figure}

Since the energy price is exogenous, we can partition the state 
$x_t=\left[
  P_t, s_t, g_t
\right]^T$
as the ``price portion" $P_t$ and ``device portion" $\left[ s_t, g_t \right]^T$.
The ``device portion" of the MDP state transition model is summarized in Figure \ref{f3}; notice that there are $(2W+1)g_{max}+\hat{W}+1$ ``device portion"
states. Recall that we use $\mathcal{P}$ to denote the set of states of the price Markov chain, so the price Markov chain has $|\mathcal{P}|$ states, and the cardinality of the state space for the device based MDP  is 
$|\State|=|\mathcal{P}| \left[ (2W+1)g_{max}+\hat{W}+1 \right]$, which is affine in $|\mathcal{P}|$, $W$, $\hat{W}$ and $g_{max}$.

Note that the action space at each state $x \in \State$ is
$\ACTION=\left \lbrace \textrm{off},\textrm{on} \right \rbrace$,
where action ``on" means the smart device completes a job\footnote{More specifically, it completes a user-initiated job if a consumer request has been received in the current episode, and completes an EMS-initiated job otherwise.} at the current time and 
``off" means the smart device does nothing at the current time.
We use $\Phi \left(x_t, a_t, x_{t+1} \right)$ to denote the instantaneous cost at
state-action-state triple $\left(x_t, a_t, x_{t+1} \right)$.

Based on the above discussion,
the proposed device based MDP model is detailed below.
\begin{itemize}
\item $P_t$ follows
 an exogenous Markov Chain with $|\mathcal{P}|$ states. 
\item If the smart device has not received a consumer request in the current episode, recall that we set
$g_t=0$ and elapsed time $s_t=t-\tau_p$. Notice that:
\begin{enumerate}
\item
If action ``off" is selected, the current cost is $\Phi \left(x_t, a_t, x_{t+1} \right)=0$. Then the smart device receives a consumer request $(t+1, \tau_g, g)$ at the next time step (time $t+1$)
with probability $p_{s_t, t+1-\tau_g, g}$, where $t+1 \le \tau_g \le t+1+W$ and $g \in \lbrace 1, 2,\cdots, g_{max} \rbrace$.
In other words, the ``device portion" state transits to $\left[ s_{t+1}=t+1-\tau_g, g_{t+1}=g \right]^T$ with probability $p_{s_t, t+1-\tau_g, g} $,
for any $t+1 \le \tau_g \le t+1+W$ and any $g \in \lbrace 1, 2,\cdots, g_{max} \rbrace$
(notice that the target time and priority together specify the type of the 
received request). On the other hand, with probability $1- \sum_{\tau_g=t+1}^{t+1+W} \sum_{g=1}^{g_{max}} p_{s_t, t+1-\tau_g, g} $, the smart device does not receive
the consumer request at time $t+1$ and 
the ``device portion" state transits to 
$\left[s_{t+1} = s_t+1, g_{t+1}=0 \right]^T $.

\item
On the other hand,
if action ``on" is selected, then the smart
device completes an EMS-initiated job at time $t$. Thus,
the current cost is $\Phi \left(x_t, a_t, x_{t+1} \right)=P_tC+\gamma \tilde{U}_e \left(s_t \right)$, where $\tilde{U}_e \left(s_t \right)$ is defined in (\ref{U_s}) and $C$ is the constant energy consumed by a standardized job. Then the current episode terminates and the smart device regenerates the ``device portion" state based on distribution $\pi_0$.
\end{enumerate}

Note that the transition probability $p_{s_t, t+1-\tau_g, g}$ and the dissatisfaction $\tilde{U}_e \left(s_t \right)$ depend on $s_t$. To ensure the 
state space is finite, we assume $p_{s_t, t+1-\tau_g, g}=p_{\hat{W}, t+1-\tau_g, g}$ and $\tilde{U}_e \left(s_t \right)=\tilde{U}_e (\hat{W} )$
for any $s_t \ge \hat{W}$.\footnote{That is, when $s_t \ge \hat{W}$
and action ``off" is selected, with probability $1-\displaystyle \sum_{\tau_g=t+1}^{t+1+W} \sum_{g=1}^{g_{max}} p_{\hat{W}, t+1-\tau_g, g} $,
the smart device will stay at the same state.}

\item If the smart device has already received a consumer request in the current episode but has not satisfied this request, recall we set 
$s_t=t-\tau_g$, where $\tau_g$ is the target time of this request.  
Notice that:
\begin{enumerate}
\item If action ``off" is selected, then with probability $\tilde{p}_{s_t, g_t}$, the ``device portion" state transits to 
$\left[ s_{t+1}= s_t+1, 
g_{t+1}=g_t \right]^T$ at time $t+1$ and 
the cost associated with this transition is $\Phi \left(x_t, a_t, x_{t+1} \right)=0$.
On the other hand,
with probability 
$1-\tilde{p}_{s_t, g_t}$, the user will cancel this request at the end of time period $t$.
In this case, the current episode terminates and the smart device regenerates the ``device portion" state based on distribution $\pi_0$ at time $t+1$. The cost associated with this transition is 
$\Phi \left(x_t, a_t, x_{t+1} \right)=\gamma \tilde{U}_c(s_t, g_t)$, where $\tilde{U}_c$ is defined in (\ref{U_c}).
Notice that 
we assume
$\tilde{p}_{s_t, g_t}=0$ if $s_t=W$.
\item If action ``on" is selected, the current cost is 
$\Phi \left(x_t, a_t, x_{t+1} \right)=P_tC+\gamma \tilde{U}_r\left(s_t, g_t \right)$, where $\tilde{U}_r\left(s_t, g_t \right)$ is defined in (\ref{U_r}) and $C$ is the constant energy consumed by a standardized job. Then the current episode 
terminates and the smart device regenerates the ``device portion" state based on distribution $\pi_0$ at time $t+1$.
\end{enumerate}
\end{itemize}

Note that if the transition model of the device based MDP and
the dissatisfaction function of the consumer are known, the device based MDP
can be solved by dynamic programming (DP). 
Following ideas in classical DP, it is straightforward to derive the Bellman equation (see Appendix A) from
which we can compute the optimal Q-function $Q^*$.
Specifically, many DP algorithms, such as value iteration and policy iteration,
can be used to compute $Q^*$ (see \cite{Bertsekas:2005}).
We observe that for many DP algorithms,
computing $Q^*$ is tractable since the 
cardinalities of the state space $\State$ and action space $\ACTION$ in a device based MDP are usually small.
Once $Q^*$ is available, one optimal policy $\mu^*$ is
$ \mu^*(x_t) \in \argmin_{a \in \ACTION} Q^* (x_t, a)$. 

Under the RL formulation, we assume that the EMS initially knows the state space $\State$, the action space $\ACTION$,
the discount $\alpha$, the trade-off parameter $\gamma$ and the per-job energy consumption $C$; but it does not 
know the state transition model or the user's dissatisfaction function. It observes the state transitions and the user's evaluations 
(which are equal to the user's dissatisfactions on rescheduling under Assumption \ref{assump:assump4})
as it interacts with the user and the real-time energy price.
At each time step, it aims to make good decisions based on its initial information, past observations and current state.

\subsection{Demand Response Potential}
\label{sec:drp}
As we have discussed in Section \ref{sec:general}, any RL algorithm can be applied to the device based MDP model proposed in Subsection \ref{sec:mdp_model}.
To justify a RL algorithm achieves satisfactory performance,
we need to compare its experimental performance with respect to the baseline policy without job rescheduling. 
In this subsection, we motivate and propose the notion of demand response potential, which upper bounds the cost reduction that can be achieved by any RL algorithm.

We start by defining some useful notation: we use $\mu: \State \times \ACTION \rightarrow [0,1]$
to denote a (randomized and stationary) policy of a device based MDP. 
Specifically, under (randomized) policy $\mu$, at any state $x =\left[P,s,g \right]^T \in \State$,
action $a \in \ACTION$ is chosen with probability $\mu(x,a)$.
As is classical in DP and RL, we use
$Q_{\mu}$ to denote the Q-function of the device based MDP under policy $\mu$ (see \cite{sutton1998reinforcement}).
Since we assume that
the energy price $P_t$ follows an exogenous ergodic Markov chain; thus, we use $\pi_P$ to denote the unique stationary
distribution of this price Markov chain. Moreover, recall that once an episode terminates, the ``device portion" state is regenerated according to
distribution $\pi_0$. Recall that
state $x=\left[P,s,g\right]^T$, thus, we use
$x \sim \pi_P \times \pi_0$ to denote $P \sim \pi_P$,
$\left[ s,g \right]^T \sim \pi_0$, and $P$ and $\left[ s,g\right]^T$ 
are statistically independent.
We define the performance of a policy $\mu$ as follows:
\BE
V_{\mu}= \mathbb{E}_{x \sim \pi_P \times \pi_0} \left \{
\mathbb{E}_{a \sim \mu \left(x , \cdot \right)} \left[
Q_\mu \left( x , a \right)
\middle |
x
\right]
\right \}. \label{V_mu}
\EE
We use $\mub$ to denote the baseline policy, and are interested in how much an RL algorithm can reduce the user's cost with respect to
$V_{\mub}$.

Recall that we use $\mu^*$ to denote the optimal policy. We define the demand response (DR) potential as
\BE
\drp = V_{\mub}-V_{\mu^*}.  \label{drp}
\EE
By definition, DR potential is the maximum expected cost reduction that can be achieved by DR.
Obviously, in the case when the transition model and the dissatisfaction function are known,
$\mu^*$ can be derived beforehand and hence $\drp$ is achieved.
However, in the practical cases when the EMS needs to learn $\mu^*$
through some RL algorithm,
$\drp$ is generally not achievable.
We define the relative DR potential ($\rdrp$) as
\BE
\rdrp= \drp/ V_{\mub} = \left( V_{\mub}-V_{\mu^*} \right)/ V_{\mub}. \label{rdrp}
\EE
Of course, $\drp$ and $\rdrp$ depend on the problem instance.
We are particularly interested in how $\drp$ and $\rdrp$ vary with $\gamma$, since $\gamma$ 
specifies 
the consumer's tradeoff between job rescheduling and the electricity bill paid. Under mild conditions, we have the following results on 
$\drp$ and $\rdrp$:
\begin{theorem}
\label{theorem1}
If (a) the energy price is always strictly positive, (b) the user's 
dissatisfaction function satisfies (\ref{U_r}-\ref{U_s}) and is always non-negative;
(c) the user's dissatisfaction is $0$ when a user's request is satisfied at its target time
or is canceled before its target time,
then we have that
\begin{enumerate}
\item $V_{\mub}$ does not depend on $\gamma$.
\item There exists a $\gamma^*>0$ s.t. 
$\mub$ is an optimal policy when $\gamma > \gamma^*$. Thus, as $\gamma \rightarrow \infty$,
$\drp \rightarrow 0$.
\item $\drp$ and $\rdrp$ are non-increasing functions of $\gamma$.
\item If $\gamma=0$, then $\rdrp=1$.
\end{enumerate}
\end{theorem}
Please refer to Appendix B for the proof of Theorem \ref{theorem1}.
Note that the condition (c) is reasonable since for a rational user, it is proper to assume his
dissatisfaction is minimal if his request is satisfied at the target time, or canceled by him before the target time.
Condition (a) and (b) can be achieved by shifting the dissatisfaction function and/or the energy price by a constant.

\subsection{Performance Metric}
\label{sec:metrics}
We now propose a performance metric for RL algorithms. We start by defining some notation: for any $t=0,1,\cdots$, use $\tilde{\mu}_t$ to denote the policy
under which the RL algorithm chooses the action $a_t$ at the beginning of time period $t$.
We define the performance of the RL algorithm as

\BE
\tilde{V}=\mathbb{E} \left[ \sum_{t=0}^{\infty} \alpha^t \Phi \left( x_t, \tilde{\mu}_t \left(x_t \right), x_{t+1} \right) \right],  \label{V_rl}
\EE
\normalsize
note that the expectation is not only taken with respect to the initial state (similarly as (\ref{V_mu}), we assume
$x_0=\left[ P_0, s_0, g_0 \right]^T \sim \pi_P \times \pi_0$
in (\ref{V_rl})), but also with respect to the subsequent stochastic transitions, noisy evaluations and possible
randomizations in the RL algorithm. Note that by definition, we have $V_{\mu^*} \le \tilde{V}$.

The performance metric of an RL algorithm is its relative improvement ($\ri$) with respect to the baseline, which is defined as
\BE
\ri=( V_{\mub}-\tilde{V}) /V_{\mub}.   \label{relative_improvement}
\EE
$\ri$ captures the normalized expected cost reduction the user will benefit from an RL algorithm.
Note that since $V_{\mu^*} \le \tilde{V}$, we have $\ri \le \rdrp$. That is, the relative DR potential is an upper bound on
the relative improvement. Notice that 
$\ri$ can be negative, since there is no guarantee that $\tilde{V} \le V_{\mub}$.
We also note that the ratio $\ri / \rdrp$  serves as an indicator on whether or not it is worthy to explicitly model the user
behavior/dissatisfaction and the energy price. 
Specifically, if $\drp$ is large but $\ri / \rdrp$ is small for many widely used RL algorithms, then 
it might be worthy to explicitly model the user behavior/dissatisfaction and the energy price, as long as the cost of such modeling is smaller than
$\tilde{V}-V_{\mu^*}$.

\subsection{Possible Extensions}
\label{sec:extension}
We now briefly discuss how to extend the device based MDP model proposed in Subsection \ref{sec:mdp_model}, by relaxing the simplifying assumptions proposed in Subsection \ref{sec:simplifying}.
\begin{itemize}
\item For device clusters consisting of more than one devices, we can propose similar MDP models with possibly higher-dimensional states.
One example is a device cluster consisting of a laundry machine and a dryer, for which we can combine these two devices as a ``super device", and still define the state as $x_t=[P_t, s_t, g_t]^T$, where $P_t$ is the energy price, $s_t$ is the elapsed time\footnote{In this example, we assume the user only cares when the dryer completes the job. Thus, there is only one target time and the elapsed time is well-defined (see (\ref{eq:elapsed_time})).}, and $g_t$ encodes information about both the \emph{stage}\footnote{In this example, the stage indicates if a request has been received,  and if the laundry machine has completed the request. Notice that we should not allow the EMS to do speculative jobs in this example.} and the priority of a job.
\item If there are other exogenous grid signals, and/or the statistical models of the grid signals and the user behavior are time-varying (e.g. periodic), then we can propose a similar MDP model by incorporating other grid signals and/or time into the state $x_t$.

\item Our proposed MDP model can be easily extended to enable the user to submit a new request while keeping the existing uncompleted requests (known as \emph{stacking requests}). One way to achieve this is to define $\tau_g$, the target time of the device, as a vector. Specifically, the cardinality of $\tau_g$ is the number of uncompleted requests to this device, and its components denote the target times of these requests. When the user submits a new request, we simply concatenate its target time to $\tau_g$.

\item In practice, the user will only evaluate some completed/canceled jobs at some later time (e.g. the end of a day). It is easy to extend our proposed MDP model to this case. One reasonable extension is to assume there is a fixed \emph{evaluation time} each day (say 9pm each day), and at the evaluation time, for each completed/canceled request in the previous $24$ hours, the user will evaluate with probability $p_{\mathrm{eval}}$, where $p_{\mathrm{eval}} \in (0,1)$ is the \emph{evaluation probability}. In general, $p_{\mathrm{eval}}$ can depend on the device, the request priority, and whether it is completed or canceled.

\end{itemize}

\section{Experiment Results}
\label{sec:experiment}
In this section, we first briefly review the classical Q-learning algorithm in Subsection \ref{sec:q_learning}.
Then we discuss a representative simulation example in Subsection
\ref{sec:setup}. Finally, the simulation results are demonstrated in 
Subsection \ref{sec:performances}.

\subsection{The Q-Learning Algorithm}
\label{sec:q_learning}

As we have discussed in Section \ref{sec:specific}, many RL algorithms can be applied to our proposed
device based RL problem. 
In this subsection, we implement one of the most popular and classical RL
algorithms, known as $Q$-learning \cite{watkins}. Q-learning is an off-policy learning algorithm; that is, it allows the learning agent to follow an exploratory policy while learning  about an optimal policy. Another desirable features of Q-learning is that it is online, incremental and is easy to implement on real-time data.

$Q$-learning works based on temporal-difference learning \cite{sutton1998reinforcement}. At each time-step $t$ the learning algorithm receives an input data in the form of $(x_{t},a_t,\Phi_t,x_{t+1})$, where 
$ \Phi_t \eqdef \Phi \left(x_t, a_t, x_{t+1} \right) $ is the observed instantaneous cost after taking action $a_t$ from state $x_t$ and arriving at state $x_{t+1}$.  Then the Q-learning algorithm updates the action-value function $Q_t(x_t,a_t)$ according to  
\begin{eqnarray}
\label{eq:qlearning}
\lefteqn{Q_{t+1}(x_t,a_t):= Q_t (x_t,a_t)} \nonumber \\
 &\,& + \beta_t \left[ \Phi_t + \alpha \min_{a' \in \ACTION} Q_t(x_{t+1},a' ) - Q_t (x_t,a_t) \right], \nonumber
\end{eqnarray}
and $Q_{t+1}(x,a):=Q_t (x,a)$ 
if $(x,a) \neq (x_t, a_t)$. Note $Q_t \in \Re^{|\mathcal{S}||\mathcal{A}|}$ is the 
state-action value function (Q-function) estimate
in time period
$t$, and $\beta_t > 0$ denotes step-size in time period $t$. 
If the step-size sequence satisfies
$\sum_{t=0}^{+\infty} \beta_t=+\infty$ and 
$\sum_{t=0}^{+\infty} \beta_t^2<+\infty$, then
Q-learning is guaranteed to converge to the optimal solution if all states are visited infinitely often (see \cite{sutton1998reinforcement}). 

To complete the description of a Q-learning algorithm, we also need to specify a behavioral policy $\mu_b$ under which
the algorithm chooses actions.
The choice of $\mu_b$ will affect data and thus would help the algorithm to learn an optimal policy faster.  One main ingredient of $\mu_b$ is that is it has to be exploratory policy during the learning process. One of standard, but crude, suggestions  for how to select $\mu_b$ is as follows: for a small $0<\epsilon<1$,  with
probability $\epsilon$ (henceforth referred to as the ``exploration probability"),  the algorithm chooses action $a_t$ from state $x_t$ according to a randomized  policy, and with probability $1-\epsilon$,  the algorithm chooses $a_t \in \argmin_{a \in \ACTION} Q_t \left(x_t, a \right)$, where $Q_t$ refers to the state-action value function estimate at time $t$. Here, we consider a randomized policy in the form of
Boltzmann exploration
\[
\pi (a_t | x_t)=
\frac{\exp \left[ -Q_t(x_t,a_t) /\eta \right]}{\sum_{a' \in \ACTION} \exp \left[ -Q_t (x_t,a')/\eta \right]} ,
\]
where $\eta>0$ is a tuning parameter and is referred to as the ``temperature" of the Boltzmann exploration.
The Q-learning algorithm that we implement in this paper is illustrated in 
Algorithm \ref{alg1}.

\begin{algorithm}[h!]
  \caption{The Q-Learning Algorithm}
  \label{alg1}
\begin{algorithmic}[1]
\STATE {\bfseries Initialize} $Q_0$ arbitrarily
\STATE {\bfseries Repeat} for each episode $j$:
\STATE Choose a small  constant step-size $\beta_j >0$ for each episode
\FOR{ each time period $t$ in episode $j$}
\STATE Take action $a_t$ at state $x_t$ according to the behavioral policy $\mu_b$ 
\STATE Observe the instantaneous cost $\Phi_t$ and new state $x_{t+1}$
\STATE Compute the TD error 
\[ \delta_t := \Phi_t +\alpha \min_{a' \in \ACTION} Q_t \left(x_{t+1},a' \right)-Q_t \left(x_t, a_t \right) \]
\STATE Update
\[
Q_{t+1}(x,a):=\left \lbrace
\begin{array}{ll}
Q_t(x,a)+\beta_j \delta_t  & \textrm{if $(x,a)=\left(x_t, a_t \right)$} \\
Q_t(x,a) & \textrm{otherwise}
\end{array}
\right. 
\]
\ENDFOR
\end{algorithmic}
\end{algorithm}

\subsection{Experiment Setup}
\label{sec:setup}

\begin{figure}
\centering
\subfigure[$\tilde{U}_r$]{
\includegraphics[width=4.5cm, height=3.4cm]{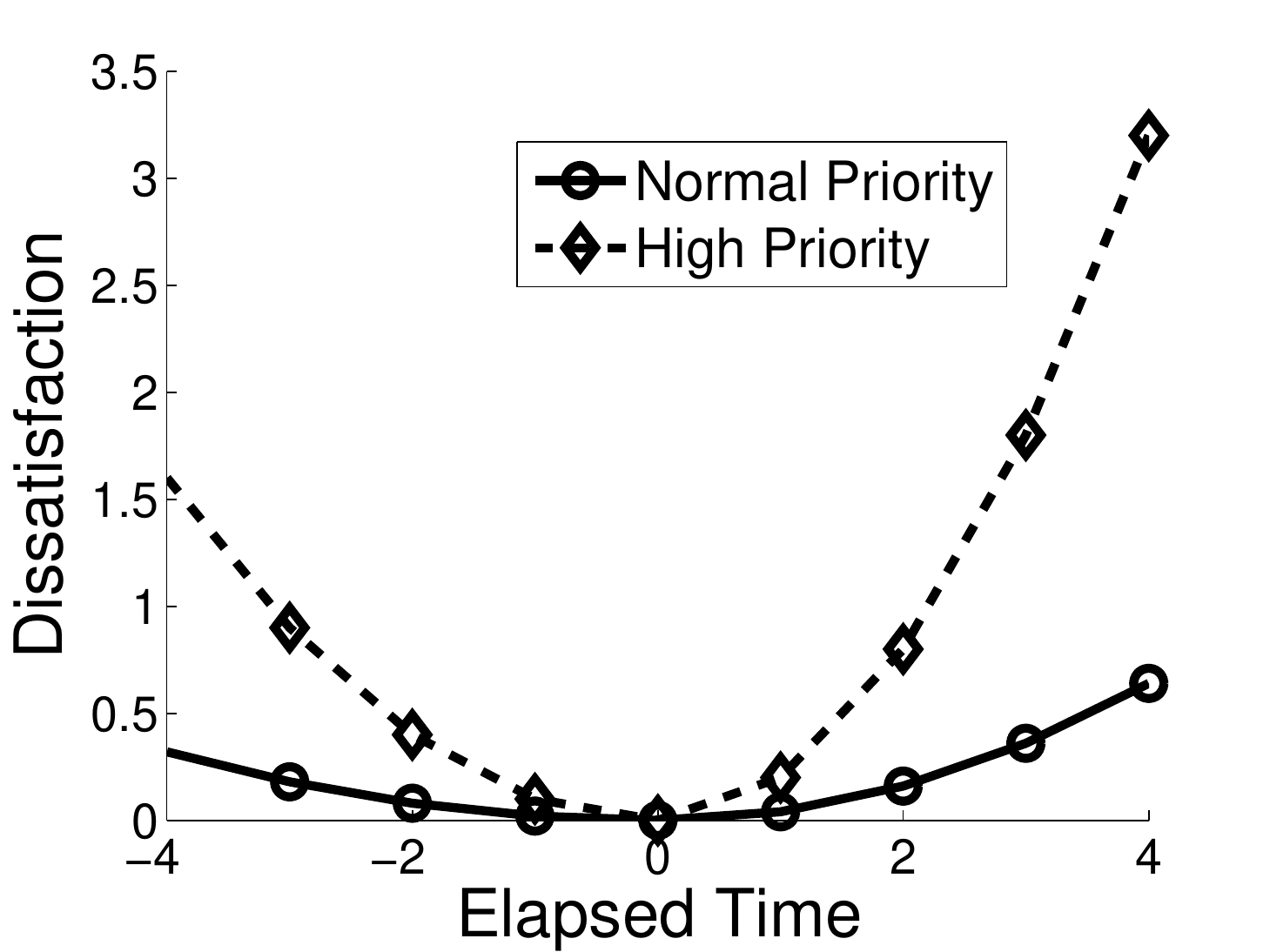}
\label{ff1:U_r}
}
\subfigure[$\tilde{U}_c$]{
\includegraphics[width=4.5cm, height=3.4cm]{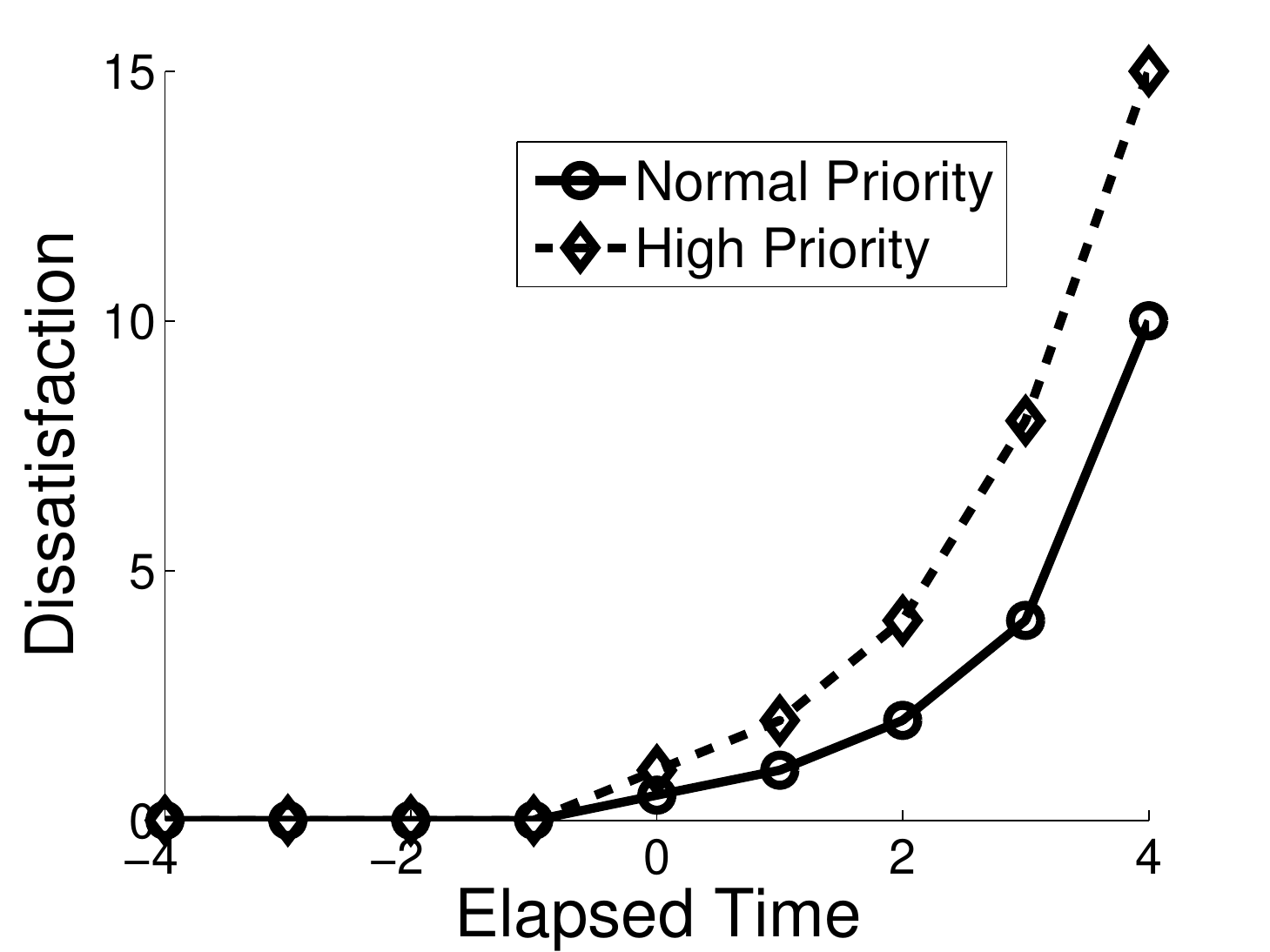}
\label{ff1:U_c}
}
\subfigure[$\tilde{U}_s$]{
\includegraphics[width=4.5cm, height=3.4cm]{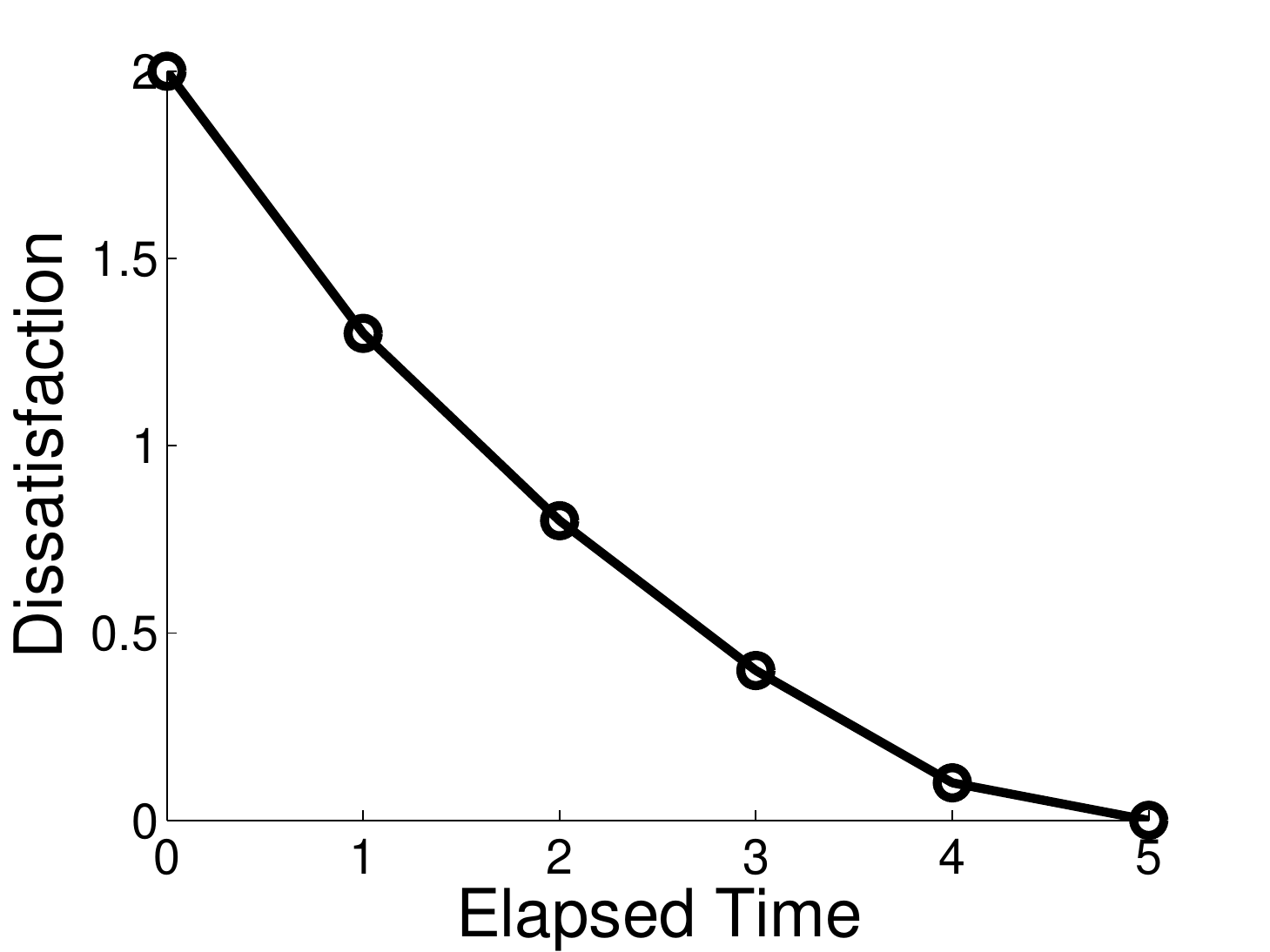}
\label{ff1:U_e}
}
\subfigure[$p_{s(t)}$]{
\includegraphics[width=4.5cm, height=3.4cm]{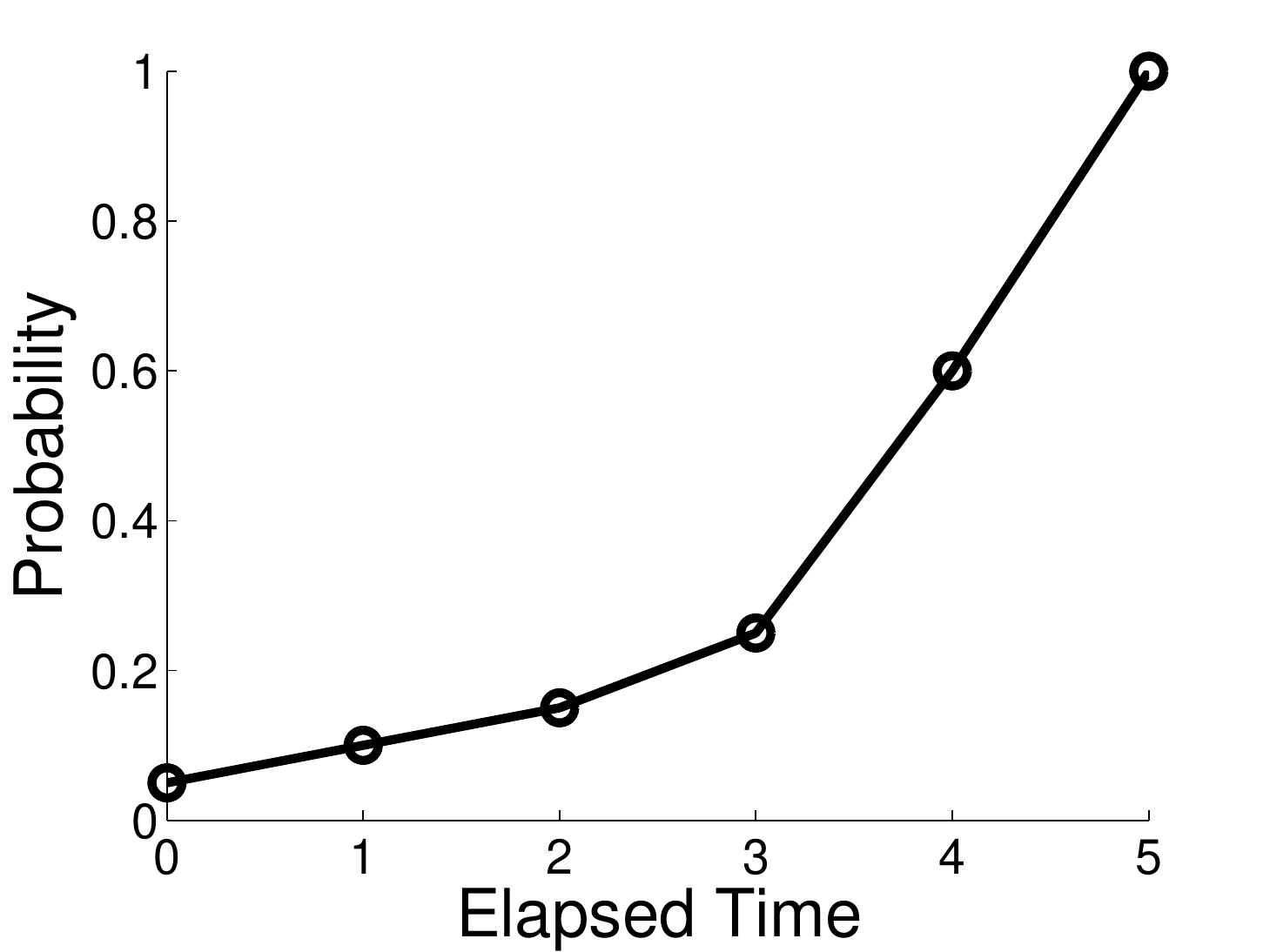}
\label{ff1:P_s}
}
\subfigure[$\hat{p}_{s(t)}$]{
\includegraphics[width=4.5cm, height=3.4cm]{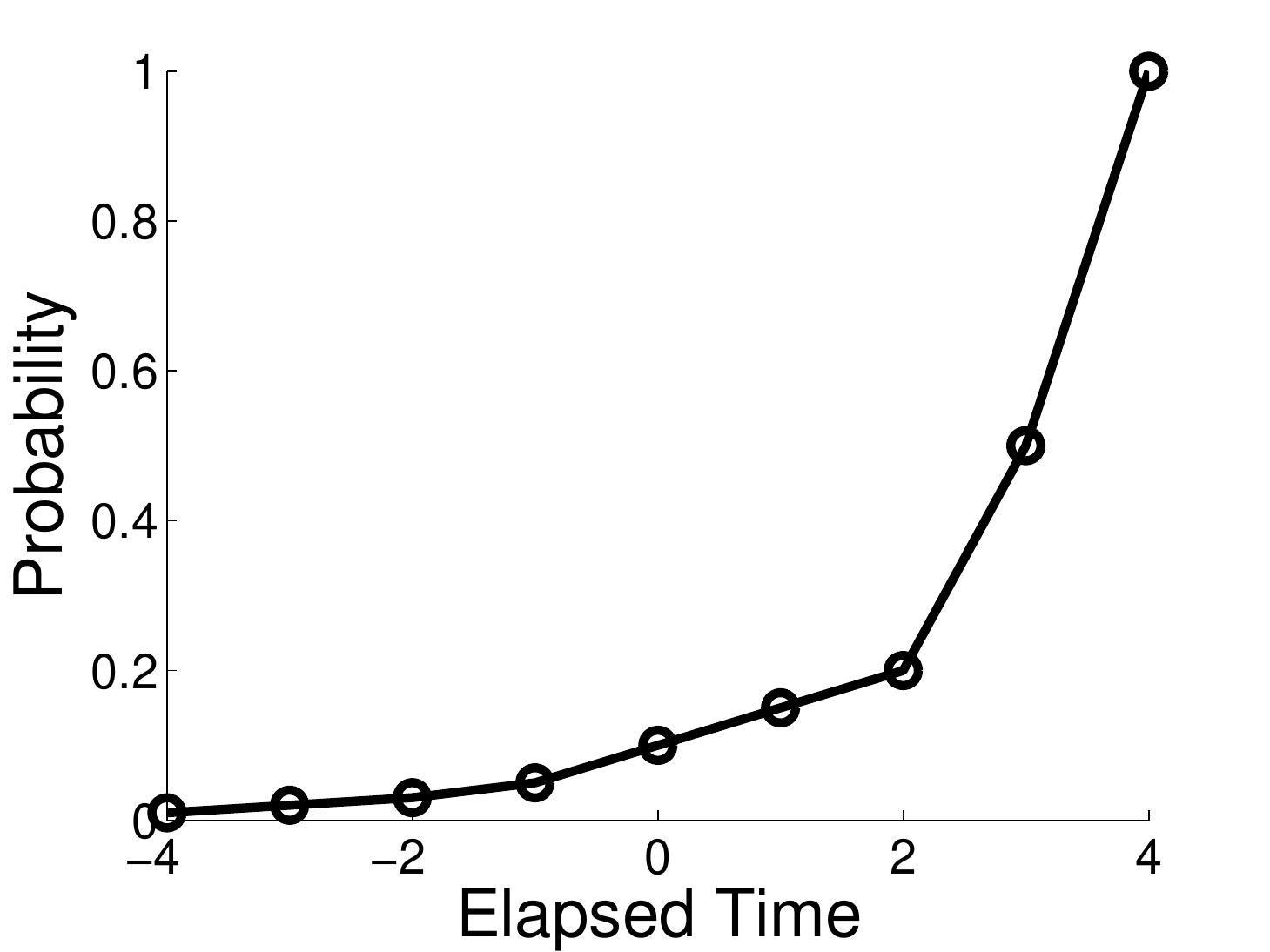}
\label{ff1:P_s_hat}
}
\label{ff1}
\caption{Dis-satisfactions Functions and Transition Model}
\end{figure}

In this subsection, we propose a representative example to which
we apply the Q-learning algorithm detailed in Algorithm \ref{alg1}.
Specifically, in this example, we assume that 
the exogenous price Markov chain
has state space $\mathcal{P}=\left \lbrace 10, 12, 15, 20 \right \rbrace$,
and the consumer requests have two different priorities, ``high" and ``normal".
We set the time windows $W=4$ and $\hat{W}=5$,
the discrete-time discount $\alpha=0.9995$ and the per job energy consumption
$C=1$.
Thus, there are $|\State|=96$ states in this example.
The dissatisfaction functions $\tilde{U}_r$, $\tilde{U}_c$ and 
$\tilde{U}_e$ are illustrated in Figure \ref{ff1:U_r}, \ref{ff1:U_c}
and \ref{ff1:U_e}. Notice that these dissatisfaction functions satisfy the 
conditions of Theorem \ref{theorem1}.

As to the transition model, we assume that if the smart device has not received a consumer request in the current episode, then under action
``off", it will receive a consumer request in the next time step with
probability $p_{s_t}$. Notice that $p_{s_t}$ is chosen to be an increasing function of the 
elapsed time $s_t$ (see Figure \ref{ff1:P_s}). 
We further notice that there are $(W+1)g_{max}=10$ types of consumer requests (with 
different target times and priorities), for simplicity, we assume these $10$ types of requests are equally likely.
Furthermore, if the smart device has received a consumer request in the current episode, then under action ``off", the unsatisfied request will be 
canceled with probability $\hat{p}_{s_t}$. In this example, we assume the ``cancellation probability" $\hat{p}$  only depends on the elapsed time $s_t$ and is 
independent of the
priority $g_t$. We choose $\hat{p}_{s_t}$  as an increasing function of $s_t$ (see Figure \ref{ff1:P_s_hat}). Finally, we assume that when the smart device regenerates its state, 
with probability
$1$, the regenerated ``device portion" state is $\left[ s_{t+1}=0, g_{t+1}=0 \right ]^T$.

\begin{figure}
\centering
\includegraphics[width=8cm,height=6cm]{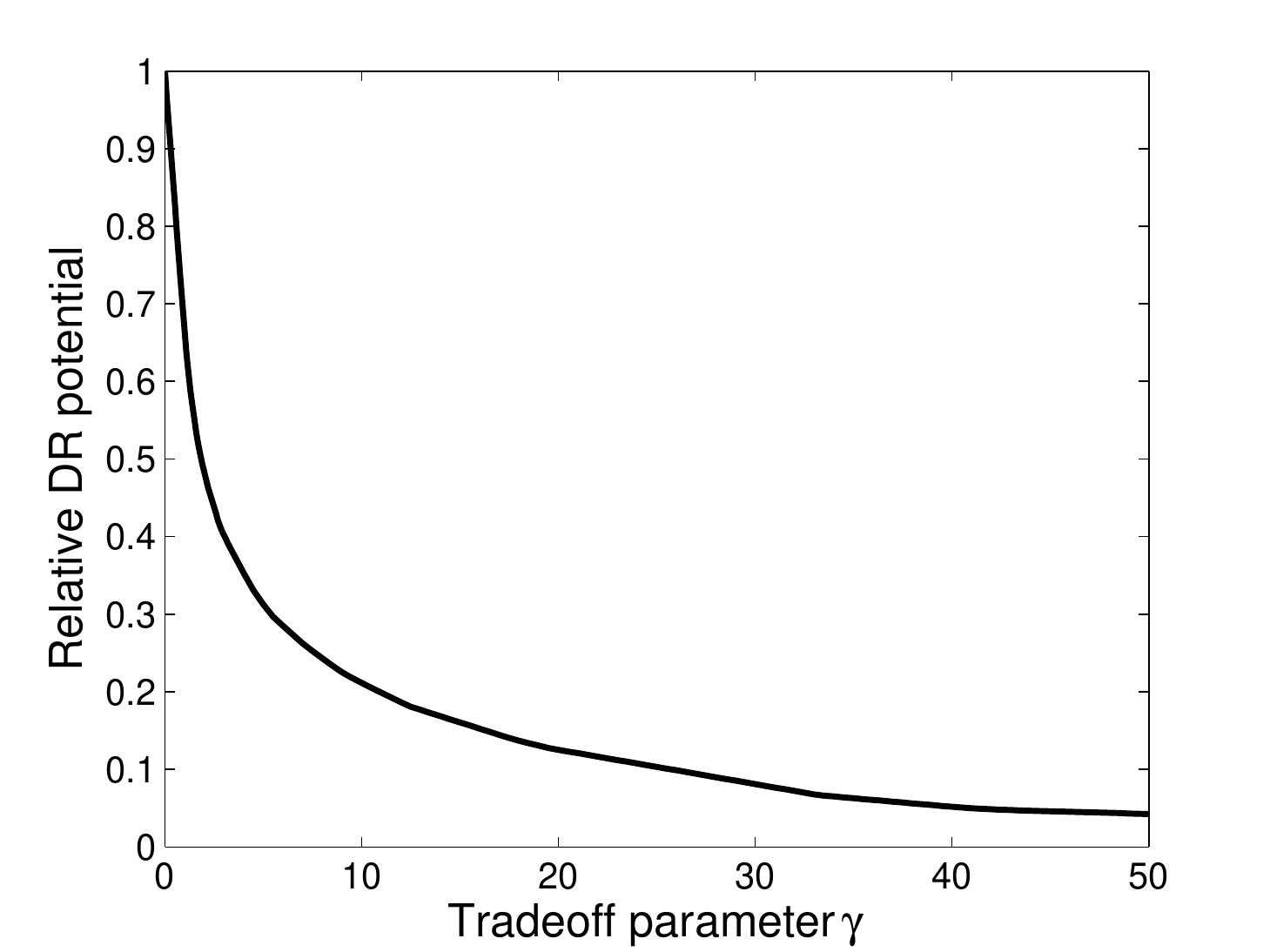}
\caption{Relative DR potential ($\rdrp$) as a Function of $\gamma$}
\label{ff2}
\end{figure}

We plot the $\rdrp$ of this example as a function of the trade-off parameter
$\gamma$ (see Figure \ref{ff2}). Since this example satisfies the conditions of Theorem \ref{theorem1}, thus, $\rdrp$ is a non-increasing function of $\gamma$. Furthermore, when $\gamma=0$,
$\rdrp=1$, and $\rdrp \rightarrow 0$ as $\gamma \rightarrow \infty$.

\subsection{Performance}
\label{sec:performances}

In this subsection, 
we present the simulation results of the Q-learning algorithm
on the representative example detailed above.
We start by describing how we implement the Q-learning algorithm.
We choose the ``exploration probability" $\epsilon=0.05$ and the
``temperature" of the softmin policy $\eta=0.1$. For episode $j$,
we choose the step-size $\beta_j=\frac{10}{20+j}$.
We initialize the Q-learning algorithm by setting
$Q_0=0$.

\begin{figure}
\centering
\includegraphics[width=8cm,height=6cm]{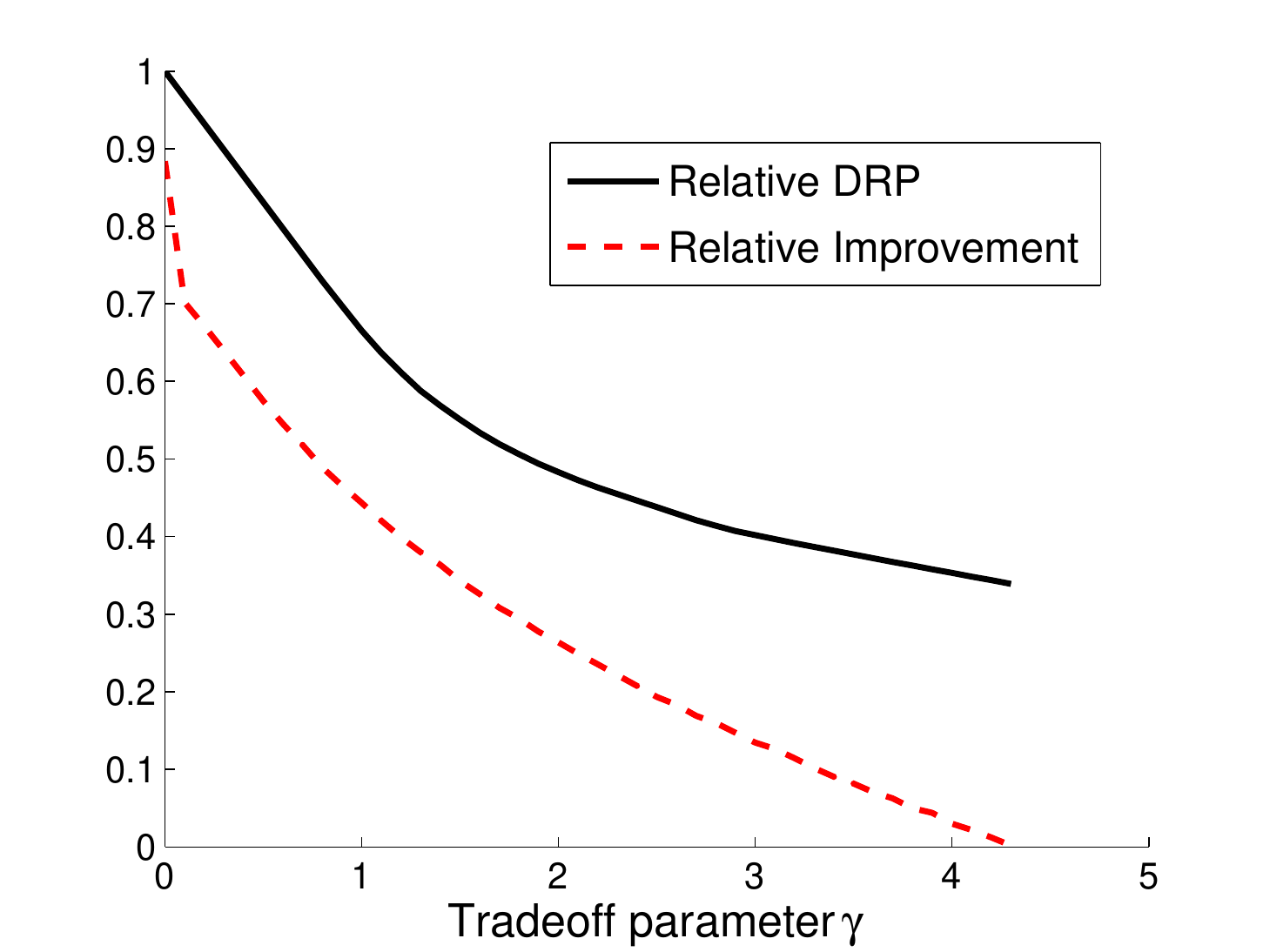}
\caption{Relative improvement ($\ri$) as a Function of $\gamma$}
\label{fig:ff3}
\end{figure}

For each trade-off parameter $\gamma=0,0.1,0.2,\cdots$, we run the
Q-learning algorithm for $\left \lceil \frac{2}{1-\alpha} \right \rceil=4,000$ episodes,
and repeat the simulation for $200$ times. Then we approximate $\tilde{V}$ by its sample mean,
i.e., by
averaging the simulation results in these $200$ simulations.
Note that $V_{\mub}$
can be analytically derived based on the Bellman equation under policy
$\mub$. Thus, $\ri$ can be (approximately) computed based on
$V_{\mub}$ and the sample mean of $\tilde{V}$.
The simulation results are summarized in Figure \ref{fig:ff3},
where we plot the $\ri$ as function of the trade-off parameter
$\gamma$. Note that we only plot the simulation results for $0 \le \gamma \le 4.3$, since for
$\gamma \ge 4.4$, we have $\tilde{V}>V_{\mub}$.
Consequently, for $\gamma \ge 4.4$, $\ri$ is negative.

We conclude this section by briefly discussing about the simulation results.
Notice that
Figure \ref{fig:ff3} shows that in this representative example,
$\ri$ is a decreasing function of $\gamma$. This implies that
with all the other parameters fixed,
the more the user prefers ``no rescheduling" to ``low energy price",
the harder for the Q-learning algorithm to achieve a significant improvement 
compared with the baseline.

\section{Conclusion and Future Work}
\label{sec:conclusion}

We have motivated and proposed a novel EMS formulation for the DR problem 
in the residential and small commercial building sectors, which we refer to as the device based RL problem. Specifically, we have shown that under appropriate assumptions,
our proposed EMS formulation does not require a pre-specified disutility function modeling the consumer's dissatisfaction
on job rescheduling, and has a computational complexity that grows linearly with the number of device clusters.
Our new EMS formulation also enables the EMS to self-initiate jobs and allows the users to the initiate more flexible requests.
We have also demonstrated the simulation results when the classical Q-learning algorithm is applied to a representative example. Simulation results suggest that
for a broad range of trade-off parameter $\gamma$, the Q-learning algorithm outperforms the baseline policy without any job rescheduling.

Finally, we briefly discuss some possible future work.
It is worth pointing out that
some assumptions in this paper (e.g. Assumption \ref{assump:assump3}) are motivated by daily observations. One possible future work is to test these assumptions based on real-world data and statistical methods.
Moreover, in this paper, we have only applied Q-learning, one of the classical RL algorithms, to a representative synthetic example.
In the future, we plan to apply some state-of-the-art RL algorithms (see \cite{2010Szepesvari,VanRoyW14}) 
to this example, as well as other synthetic/real-world examples.

\begin{center}
\textbf{Acknowledgment}
\end{center}

We acknowledge Prof. Benjamin Van Roy for his insightful comments on this paper. 

\bibliographystyle{plain}
\bibliography{IEEEabrv,lit_review,IEEEexample}

\begin{thebibliography}{25}
\providecommand{\natexlab}[1]{#1}
\providecommand{\url}[1]{\texttt{#1}}
\expandafter\ifx\csname urlstyle\endcsname\relax
  \providecommand{\doi}[1]{doi: #1}\else
  \providecommand{\doi}{doi: \begingroup \urlstyle{rm}\Url}\fi

\bibitem[Barbose et~al.(2004)Barbose, Goldman, and Neenan]{barbose2004survey}
G.~Barbose, C.~Goldman, and B.~Neenan.
\newblock {A survey of utility experience with real time pricing}.
\newblock \emph{Lawrence Berkeley National Laboratory: Lawrence Berkeley
  National Laboratory}, 2004.

\bibitem[Bertsekas(2005)]{Bertsekas:2005}
D.~Bertsekas.
\newblock \emph{Dynamic Programming and Optimal Control}.
\newblock Athena Scientific, Massachusetts, 2005.

\bibitem[Bertsekas and Tsitsiklis(1996)]{Bertsekas1996}
Dimitri~P. Bertsekas and John Tsitsiklis.
\newblock \emph{Neuro-Dynamic Programming}.
\newblock Athena Scientific, September 1996.

\bibitem[Borenstein et~al.(2002)Borenstein, Jaske, and
  Rosenfeld]{borenstein2002dynamic}
S.~Borenstein, M.~Jaske, and A.~Rosenfeld.
\newblock Dynamic pricing, advanced metering, and demand response in
  electricity markets.
\newblock \emph{UC Berkeley: Center for the Study of Energy Markets}, October
  2002.

\bibitem[Braithwait and Eakin(2002)]{braithwait2002role}
S~Braithwait and Kelly Eakin.
\newblock The role of demand response in electric power market design.
\newblock \emph{Edison Electric Institute}, 2002.

\bibitem[Coates et~al.(2010)Coates, Abbeel, and Ng]{CoatesAN10}
Adam Coates, Pieter Abbeel, and Andrew~Y. Ng.
\newblock Autonomous helicopter flight using reinforcement learning.
\newblock In \emph{Encyclopedia of Machine Learning}, pages 53--61. 2010.

\bibitem[Faruqui and George(2005)]{Faruqui200553}
Ahmad Faruqui and Stephen George.
\newblock Quantifying customer response to dynamic pricing.
\newblock \emph{The Electricity Journal}, 18\penalty0 (4):\penalty0 53 -- 63,
  2005.

\bibitem[Herter(2007{\natexlab{a}})]{Herter20072121}
Karen Herter.
\newblock Residential implementation of critical-peak pricing of electricity.
\newblock \emph{Energy Policy}, 35\penalty0 (4):\penalty0 2121 -- 2130,
  2007{\natexlab{a}}.
\newblock ISSN 0301-4215.

\bibitem[Herter(2007{\natexlab{b}})]{karen}
Karen Herter.
\newblock {An exploratory analysis of California residential customer response
  to critical peak pricing of electricity}.
\newblock \emph{Energy}, 32\penalty0 (1):\penalty0 25 -- 34, January
  2007{\natexlab{b}}.

\bibitem[Kara et~al.(2012)Kara, Berges, Krogh, and Kar]{kara2012}
Emre~Can Kara, Mario Berges, Bruce Krogh, and Soummya Kar.
\newblock Using smart devices for system-level management and control in the
  smart grid: A reinforcement learning framework.
\newblock In \emph{Smart Grid Communications (SmartGridComm), 2012 IEEE Third
  International Conference on}, pages 85--90. IEEE, 2012.

\bibitem[Koch and Piette(2007)]{kochautomated}
E.~Koch and M.A. Piette.
\newblock {Architecture concepts and technical issues for an open,
  interoperable automated demand response infrastructure}.
\newblock In \emph{Grid Interop Forum}, {Albuquerque, NM, US}, November 2007.

\bibitem[LeMay et~al.(2008)LeMay, Nelli, Gross, and Gunter]{lemay2008}
M.~LeMay, R.~Nelli, G.~Gross, and C.~A. Gunter.
\newblock {An integrated architecture for demand response communications and
  control}.
\newblock \emph{in Proc. of the 41st Hawaii International Conference on System
  Sciences}, 2008.

\bibitem[ONeill et~al.(2010)ONeill, Levorato, Goldsmith, and Mitra]{oneill2010}
D.~ONeill, M.~Levorato, A.~J. Goldsmith, and U.~Mitra.
\newblock Residential demand response using reinforcement learning.
\newblock In \emph{IEEE SmartGridComm}, {Gaithersburg, Maryland, USA}, OCT
  2010.

\bibitem[Piette et~al.(2005)Piette, Sezgen, D.Watson, Motegi, Shockman, and ten
  Hope]{report}
M.~A. Piette, O.~Sezgen, D.Watson, N.~Motegi, C.~Shockman, and L.~ten Hope.
\newblock Development and evaluation of fully automated demand response in
  large facilities.
\newblock January 2005.

\bibitem[Piette et~al.(2007)Piette, D.Watson, Motegi, and
  Kiliccote]{piettefield}
M.~A. Piette, D.Watson, N.~Motegi, and S.~Kiliccote.
\newblock {Automated critical peak pricing field tests: 2006 pilot program
  description and results}.
\newblock In \emph{LBNL Report 62218}, {Albuquerque, NM, US}, May 2007.

\bibitem[Roos and Lane(1998)]{651628}
J.G. Roos and I.E. Lane.
\newblock Industrial power demand response analysis for one-part real-time
  pricing.
\newblock \emph{Power Systems, IEEE Transactions on}, 13\penalty0 (1):\penalty0
  159 --164, feb 1998.

\bibitem[Sutton and Barto(1998)]{sutton1998reinforcement}
R.S. Sutton and A.G. Barto.
\newblock \emph{Reinforcement learning}.
\newblock MIT Press, 1998.

\bibitem[Szepesv{\'a}ri(2010)]{2010Szepesvari}
Csaba Szepesv{\'a}ri.
\newblock \emph{Algorithms for Reinforcement Learning}.
\newblock Synthesis Lectures on Artificial Intelligence and Machine Learning.
  Morgan {\&} Claypool Publishers, 2010.

\bibitem[Turitsyn et~al.(2011)Turitsyn, Backhaus, Ananyev, and
  Chertkov]{Turitsyn2011}
Konstantin~S. Turitsyn, Scott Backhaus, Maxim Ananyev, and Michael Chertkov.
\newblock Smart finite state devices: A modeling framework for demand response
  technologies.
\newblock \emph{CoRR}, abs/1103.2750, 2011.

\bibitem[Van~Roy and Wen(2014)]{VanRoyW14}
Benjamin Van~Roy and Zheng Wen.
\newblock Generalization and exploration via randomized value functions.
\newblock \emph{CoRR}, abs/1402.0635, 2014.

\bibitem[Watkins(1989)]{watkins}
C.~Watkins.
\newblock \emph{Learning from Delayed Rewards}.
\newblock PhD thesis, University of Cambridge, 1989.

\bibitem[Wen(2014)]{wen2014dissertation}
Zheng Wen.
\newblock \emph{Efficient Reinforcement Learning with Value Function
  Generalization}.
\newblock PhD thesis, Stanford University, 2014.

\bibitem[Wen and Van~Roy(2013)]{WenVR13_deterministic}
Zheng Wen and Benjamin Van~Roy.
\newblock Efficient exploration and value function generalization in
  deterministic systems.
\newblock In \emph{NIPS}, pages 3021--3029, 2013.

\bibitem[Wen et~al.(2011)Wen, Durlofsky, Van~Roy, and Aziz]{WenDVA2011}
Zheng Wen, Louis~J. Durlofsky, Benjamin Van~Roy, and Khalid Aziz.
\newblock Use of approximate dynamic programming for production optimization.
\newblock In \emph{SPE Proceedings}, {the Woodlands, Texas, USA}, February
  2011.

\bibitem[Wen et~al.(2012)Wen, Durlofsky, Van~Roy, and Aziz]{WenDVA2012}
Zheng Wen, Louis~J. Durlofsky, Benjamin Van~Roy, and Khalid Aziz.
\newblock Approximate dynamic programming for optimizing oil production.
\newblock In Frank~L. Lewis and Derong Liu, editors, \emph{Reinforcement
  Learning and Approximate Dynamic Programming for Feedback Control}.
  Wiley-IEEE Press, 2012.

\end{thebibliography}

\newpage
\appendix
\section{Bellman Equation}
It is worth pointing out that if the transition model of the device based MDP and
the dissatisfaction function of the consumer are known, the device based MDP
can be solved by dynamic programming (DP). 
Following ideas in classical DP,
in this section, we derive the Bellman equation from which we can compute the optimal
Q-function.\\
\\
\noindent
Recall that $x_t=\left[ P_t, s_t, g_t \right]^T$ and the action space at each state is $\ACTION=\left \lbrace \textrm{off}, \textrm{on} \right \rbrace$,
we have
\begin{itemize}
\item If the smart device has not received a consumer request in the current episode, we have
\small
\BE
Q^*\left(x_t, \textrm{on} \right) =
P_tC+\gamma \tilde{U}_e \left(s_t \right)
+
\alpha \mathbb{E}\left \lbrace
\min_{a \in \ACTION}Q^*\left( \left[ P_{t+1}, s_{t+1}, g_{t+1} \right]^T, a \right)
\right \rbrace
, \nonumber
\EE
\normalsize
where the expectation is over $P_{t+1}$, $s_{t+1}$ and $g_{t+1}$.
Specifically, $P_{t+1}$ is drawn according to the transition probability of the price Markov chain, and 
$[s_{t+1},g_{t+1}]^T$ is drawn according to the ``device portion" regeneration distribution $\pi_0$.\\
\\
On the other hand, we have 
\small
\begin{eqnarray*}
Q^*\left(x_t, \textrm{off} \right) &=&
\alpha \mathbb{E}
\left[ \sum_{s'=-W}^{0} \sum_{g=1}^{g_{max}} p_{s_t, s', g} \min_{a \in \ACTION} Q^* \left( \left[P_{t+1}, s', g \right]^T, a \right) \right. \nonumber \\
&+& \left. \left(1-p_{s_t} \right) \min_{a \in \ACTION} Q^* \left( \left[P_{t+1}, \tilde{s}(t+1), 0 \right]^T, a \right)
\right],  \nonumber
\end{eqnarray*}
\normalsize
where the expectation is over $P_{t+1}$.
Specifically,
$p_{s_t}=\sum_{s'=-W}^{0} \sum_{g=1}^{g_{max}} p_{s_t, s', g}$ is the probability that a consumer request will be received in the
next time step, $\tilde{s}(t+1)=s_t+1$ if $s_t<\hat{W}$ and $\tilde{s}(t+1)=\hat{W}$ if $s_t=\hat{W}$, and 
$P_{t+1}$ is drawn according to the transition probability of the price Markov chain.\\

\item If the smart device has already received a consumer request in the current episode, we have
\BE
Q^*\left(x_t, \textrm{on} \right) =
P_tC+\gamma \tilde{U}_r \left(s_t, g_t \right) 
+
\alpha \mathbb{E}\left \lbrace
\min_{a \in \ACTION}Q^*\left( \left[ P_{t+1}, s_{t+1}, g_{t+1} \right]^T, a \right)
\right \rbrace,  \nonumber
\EE
\normalsize
where the expectation is over $P_{t+1}$, $s_{t+1}$ and $g_{t+1}$.
Similarly, $P_{t+1}$ is drawn according to the transition probability of the price Markov chain, and 
$[s_{t+1},g_{t+1}]^T$ is drawn according to the ``device portion" regeneration distribution $\pi_0$.\\

On the other hand, we have
\begin{eqnarray*}
Q^*\left(x_t, \textrm{off} \right) &=& \alpha \tilde{p}_{s_t, g_t} \mathbb{E}
\left[
\displaystyle \min_{a \in \ACTION} Q^* \left( \left[P_{t+1}, s_t+1, g_t \right]^T, a \right)
\right] 
+ (1-\tilde{p}_{s_t, g_t}) \left[ \gamma \tilde{U}_c \left(s_t, g_t \right) \right.  \nonumber \\
&+& 
\left.
\alpha \mathbb{E}\left \lbrace
\min_{a \in \ACTION}Q^*\left( \left[ P_{t+1}, s_{t+1}, g_{t+1} \right]^T, a \right)
\right \rbrace
\right].
 \nonumber 
\end{eqnarray*}
\normalsize

Note that $\tilde{p}_{s_t, g_t}=0$ if $s_t=W$.
The expectation in the first line is over $P_{t+1}$, where 
$P_{t+1}$ is drawn according to the transition probability of the price Markov chain.
On the other hand,
the expectation in the third line is over $P_{t+1}$, $s_{t+1}$ and $g_{t+1}$, where
$P_{t+1}$ is also drawn according to the transition probability of the price Markov chain, and 
$[s_{t+1},g_{t+1}]^T$ is drawn according to the ``device portion" regeneration distribution $\pi_0$.
\end{itemize}
From the classical DP theory, the optimal Q-function $Q^*$
is the unique solution of the above-derived Bellman equation.
Furthermore, many DP algorithms, such as value iteration and policy iteration,
can be used to compute $Q^*$.
We observe that for many DP algorithms,
computing $Q^*$ is tractable since the 
cardinalities of the state space $\State$ and action space $\ACTION$ in a device based MDP are usually small.
Once $Q^*$ is available, one optimal policy $\mu^*$ is
\[ \mu^*(x_t) \in \argmin_{a \in \ACTION} Q^* (x_t, a) . \]

\section{Proof for Theorem 1}
\noindent
\textbf{Proof for Theorem 1:}\\
Note that $\forall \mu: \State \times \ACTION \rightarrow [0,1]$, $V_{\mu}$ can be expressed as
\[
V_{\mu}=A_{\mu}+\gamma  B_{\mu},
\]
where $A_{\mu}$ is the expected infinite-horizon discounted electricity bill, and
$B_{\mu}$ is the expected infinite-horizon discounted user dissatisfaction on rescheduling.
Furthermore, neither $A_{\mu}$ nor $B_{\mu}$ depends on $\gamma$.

First, we prove that $V_{\mub}$ does not depend on $\gamma$.
Note that under policy $\mub$, the EMS will never initiate a job; 
furthermore, any job initiated by the user is either completed at its target time
or canceled by the user before its target time. Thus, under Assumption 5,
we have $B_{\mub}=0$. So we have $V_{\mub}=A_{\mub}$, which does not depend on 
$\gamma$.

Second, we prove that there exists a $\gamma^*>0$ s.t. 
$\mub$ is an optimal policy when $\gamma > \gamma^*$.
Note that under Assumption 5, $\mub$, which is a deterministic policy, minimizes $B_{\mu}$, since
$B_{\mu} \ge 0$ for any $\mu$ and $B_{\mub}=0$. Note that there are finite deterministic policies,
and we use $\Delta B$ to denote the ``second best" $B_{\mu}$ under deterministic policies
\[
\Delta B=\min_{\textrm{deterministic }\mu: B_{\mu}>0} B_{\mu}.
\]
Note that the maximum cost reduction from $A_{\mu}$ is $\frac{1}{1-\alpha} \left( P_{\max}-P_{\min} \right)C$,
where $P_{\max}=\max_{P \in \mathcal{P}} P$ is the highest energy price and 
$P_{\min}=\min_{P \in \mathcal{P}} P$ is the lowest energy price.
Thus, if we choose
\[
\gamma^*=\frac{\left( P_{\max}-P_{\min} \right)C}{(1-\alpha) \Delta B},
\]
then, $\forall \gamma > \gamma^*$, $\mub$ outperforms all the deterministic policies (i.e. $V_{\mub} \le V_{\mu}$ for any deterministic $\mu$).
Consequently, it is an optimal policy. Thus $\drp=0$ for any $\gamma> \gamma^*$.
As a result, we have $\drp \rightarrow 0$ as $\gamma \rightarrow \infty$.

Third, we prove that $\drp$ is a non-increasing function of $\gamma$.
To formalize the result, we use
$V_{\mu}(\gamma)$ to denote the performance of policy $\mu$ at tradeoff parameter $\gamma$.
Recall $V_{\mu}(\gamma)=A_{\mu}+\gamma  B_{\mu}$, and $B_{\mu} \ge 0$ under Assumption 5, thus
$V_{\mu}(\gamma)$ is a non-decreasing function of $\gamma$ for any $\mu$.
Furthermore, we use $\mu^*(\gamma)$ to denote an optimal policy in the problem instance with tradeoff parameter
$\gamma$. Thus,
for any $0 \le \gamma_1 \le \gamma_2$, we have that
\[
V_{\mu^*(\gamma_1)}(\gamma_1) \le V_{\mu^*(\gamma_2)}(\gamma_1),
\]
since $\mu^*(\gamma_1)$ is an optimal policy in the problem instance with parameter $\gamma_1$, and
\[
V_{\mu^*(\gamma_2)}(\gamma_1) \le V_{\mu^*(\gamma_2)}(\gamma_2),
\]
since $\mu^*(\gamma_2)$ is a fixed policy and $\gamma_1 \le \gamma_2$.
Thus, $V_{\mu^*(\gamma_1)}(\gamma_1) \le V_{\mu^*(\gamma_2)}(\gamma_2) $, that is,
$V_{\mu^*}$ is a non-decreasing function of $\gamma$.
Since $\drp=V_{\mub}-V_{\mu^*}$, and $V_{\mub}$ does not depend on $\gamma$, and 
$V_{\mu^*}$ is non-decreasing in $\gamma$, thus, $\drp$ is non-increasing in $\gamma$.
From the definition of $\rdrp$, it is also non-increasing in $\gamma$.

Finally, we prove that $\rdrp=1$ when $\gamma=0$. 
Note when $\gamma=0$, the 
EMS does not care about the user's dis-satisfaction on job rescheduling. 
Consider  a policy $\mu'$ under which 
\begin{itemize}
\item The EMS will never initiate a job.
\item The EMS ignores all the jobs initiated by the user; it just waits for the user to cancel the requested job.
\end{itemize}
Obviously, when $\gamma=0$, we have $V_{\mu'}=0$. Thus we have $V_{\mu^*} \le V_{\mu'}=0$. On the other hand, from Assumption 5,
we have $V_{\mu^*} \ge 0$ (since the energy price is always strictly positive and and the user's dissatisfaction function is always non-negative). Thus we have $V_{\mu^*}=0$ and 
\[
\rdrp=\frac{V_{\mub}-V_{\mu^*}}{V_{\mub}}=1.
\]
Q.E.D.
\end{document}